\documentclass{article}
\usepackage{ijcai24}

\usepackage{times}
\usepackage{soul}
\usepackage{url}
\usepackage[hidelinks]{hyperref}
\usepackage[utf8]{inputenc}
\usepackage[small]{caption}
\usepackage{graphicx}
\usepackage{amsmath}
\usepackage{amsthm}
\usepackage{booktabs}
\usepackage{algorithm}
\usepackage{algorithmic}
\usepackage{colortbl}
\usepackage[switch]{lineno}
\usepackage{amsthm,amsmath,amssymb,amsfonts,bm}
\usepackage{subcaption}
\usepackage{xcolor}
\usepackage{textcomp}
\usepackage{footmisc}

\def\eg{\emph{e.g}.} 
\def\ie{\emph{i.e}.}

\def\method{RNA}
\definecolor{LightCyan}{rgb}{0.88, 0.96, 0.92}
\definecolor{harmonic}{HTML}{00A74F}
\definecolor{inharmonic}{HTML}{EA9424}
\definecolor{source}{HTML}{929292}

\newcommand{\blfootnote}[1]{
  \begingroup
  \renewcommand\thefootnote{}\footnote{#1}
  \addtocounter{footnote}{-1}
  \endgroup
}

\newcommand{\paratitle}[1]{\vspace{1ex}\noindent\emph{\textbf{#1}}}
\usepackage{amsmath,amsfonts,bm}
\usepackage{algorithm,amsmath,amssymb,bm,amsthm}
\usepackage{algorithmic}
\usepackage{enumitem}
\newtheorem{theorem}{Theorem}[section]

\newtheorem*{lemma*}{Lemma}

\def\eqref#1{equation~\ref{#1}}
\def\1{\bm{1}}

\def\rvc{{\mathbf{c}}}

\def\rve{{\mathbf{e}}}

\def\rvh{{\mathbf{h}}}

\def\rvz{{\mathbf{z}}}

\def\rmL{{\mathbf{L}}}

\def\mA{{\bm{A}}}
\def\mB{{\bm{B}}}

\def\mL{{\bm{L}}}

\def\mR{{\bm{R}}}
\def\mS{{\bm{S}}}

\def\mU{{\bm{U}}}

\DeclareMathAlphabet{\mathsfit}{\encodingdefault}{\sfdefault}{m}{sl}
\SetMathAlphabet{\mathsfit}{bold}{\encodingdefault}{\sfdefault}{bx}{n}

\def\gD{{\mathcal{D}}}

\def\sC{{\mathbb{C}}}

\def\sH{{\mathbb{H}}}
\def\sI{{\mathbb{I}}}

\DeclareMathOperator*{\argmin}{arg\,min}

\title{Rank and Align: Towards Effective Source-free Graph Domain Adaptation}

\author{
Junyu Luo$^1$
\and
Zhiping Xiao$^2$\textsuperscript{\textdagger} \and
Yifan Wang$^3$\and
Xiao Luo$^2$\textsuperscript{\textdagger} \and
Jingyang Yuan$^1$ \and \\
Wei Ju$^1$ \and
Langechuan Liu$^4$ \And
Ming Zhang$^1$\textsuperscript{\textdagger} \\
\affiliations
$^1$ School of Computer Science, National Key Laboratory for Multimedia Information Processing \\ Peking University-Anker Embodied AI Lab, Peking University, Beijing\\
$^2$ University of California, Los Angeles\\
$^3$ University of International Business and Economics \\
$^4$ Anker Innovations
}
\begin{document}

\maketitle

\blfootnote{
\textsuperscript{\textdagger} Corresponding authors.
}

\begin{abstract}

Graph neural networks (GNNs) have achieved impressive performance in graph domain adaptation. However, extensive source graphs could be unavailable in real-world scenarios due to privacy and storage concerns. To this end, we investigate an underexplored yet practical problem of source-free graph domain adaptation, which transfers knowledge from source models instead of source graphs to a target domain. To solve this problem, we introduce a novel GNN-based approach called \underline{R}ank a\underline{n}d \underline{A}lign~(\method{}), which ranks graph similarities with spectral seriation for robust semantics learning, and aligns inharmonic graphs with harmonic graphs which close to the source domain for subgraph extraction. In particular, to overcome label scarcity, we employ the spectral seriation algorithm to infer the robust pairwise rankings, which can guide semantic learning using a similarity learning objective. To depict distribution shifts, we utilize spectral clustering and the silhouette coefficient to detect harmonic graphs, which the source model can easily classify. To reduce potential domain discrepancy, we extract domain-invariant subgraphs from inharmonic graphs by an adversarial edge sampling process, which guides the invariant learning of GNNs. Extensive experiments on several benchmark datasets demonstrate the effectiveness of our proposed \method{}. 

\end{abstract}

\section{Introduction}

\begin{figure}[t]
    \centering
    \includegraphics[width=\linewidth]{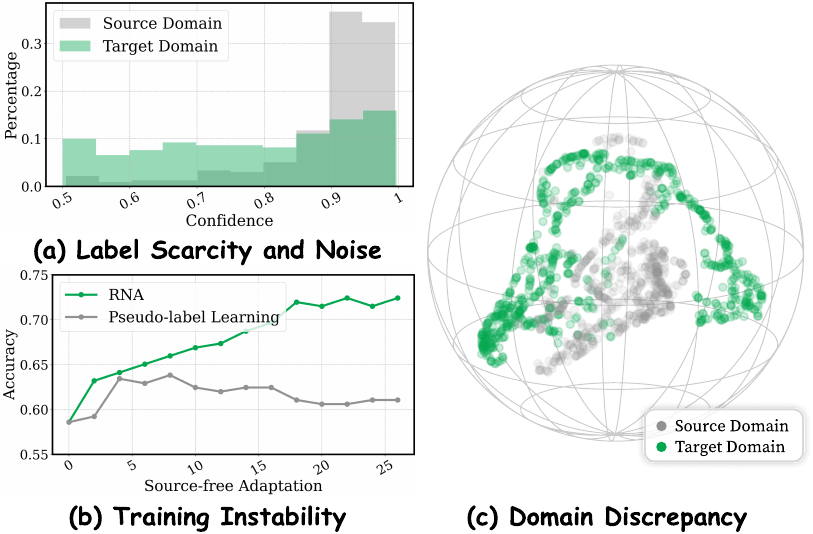}
    \caption{ Motivation of \method{}.
    \textbf{(a)} The histogram of confidence score of the source-trained model perform on \textcolor{source}{source} and \textcolor{harmonic}{target} data, highlighting the label scarcity and potential noise in the target domain. 
    \textbf{(b)} The training process of \method{} and adaptation with pseudo-label learning.
    \textbf{(c)} Domain discrepancy between \textcolor{source}{source} and \textcolor{harmonic}{target} domain.
    Best viewed in color and zoom-in.
    }
    \label{fig:fig1}
\end{figure}

Graph Neural Networks~(GNNs) have achieved great success in a wide range of applications, including molecular generation~\cite{kim2023learning}, traffic networks~\cite{kdd-traffic1}, social networks~\cite{application_1,application_2} and relational databases~\cite{sql-gnn-1}. They could be used to solve the graph classification problem, which aims to predict the labels of entire graphs~\cite{zhang2018end,ying2018hierarchical,wu2020comprehensive}. Most of these approaches adopt the message-passing paradigm~\cite{GCN} to update node-level representation iteratively, followed by a readout operator to summarize the node-level representations to graph-level representations for downstream classifications. 

Despite their superior performance, these approaches usually assume that the training and test data are from the same data distribution, which is often not the case in real-world scenarios~\cite{uda-gcn,you2022graph,lin2023multi}. The out-of-distribution challenge has promoted the development of unsupervised graph domain adaptation~\cite{hao2020asgn,coco}. However, these approaches require access to a label-rich source domain to adapt to the label-scarce target domain, meaning that abundant source graphs are needed. The requirement could be difficult due to privacy and storage concerns in practice. This motivates us to study a practical yet underexplored research problem of source-free graph domain adaptation. The objective is to transfer these pre-trained graph models from the source domain to the target domain without accessing source data.  

However, formalizing an effective framework for source-free graph domain adaptation is a non-trivial problem, which requires us to solve the following two research problems. 
\textit{(1) How to learn semantics on the target domain, given its label scarcity?} Previous approaches~\cite{ding2022source,zhang2022divide} usually utilize pseudo-labeling to learn from target data. However, serious distribution shifts could generate biased and inaccurate pseudo-labels, which results in extensive error accumulation and training instability, as demonstrated in Figure~\ref{fig:fig1}(a)~(b). \textit{(2) How to deal with extensive distribution shifts without access to source data?} 
Source and target graphs could belong to very different domains,
which strongly decreases the accuracy of predictions. Previous graph domain adaptation approaches~\cite{yin2022deal,coco} usually minimize the distribution discrepancy between the source and target domains, which is not feasible due to the requirement of extensive source data. 

In this paper, we propose a novel approach named \underline{R}ank a\underline{n}d \underline{A}lign (\method{}) for source-free graph domain adaptation. The high-level idea of our \method{} is to infer spectral seriation rankings to guide robust semantics learning and detect harmonic graphs (\ie, close to source graphs) for domain alignment. In particular, \method{} utilizes the spectral algorithm to generate seriation similarity rankings among target graphs, which are robust to the potential noise. Then, a ranking-based similarity learning objective is adopted to guide the semantics learning under the target label scarcity. To depict the distribution shift, we introduce the silhouette coefficient after spectral clustering, which identifies harmonic graphs closely related to source graphs on the target domain. To further reduce potential domain discrepancy, we extract subgraphs from inharmonic graphs by sampling edges with adversarial learning, which discards information irrelevant to semantics labels. Then, we conduct invariant learning to ensure our GNNs are insensitive to these irrelevant parts. Finally, we utilize pseudo-labels with multi-view filtering to enhance semantics learning with alleviated error accumulation. Extensive experiments on several benchmark datasets validate the superiority of \method{} in comparison to extensive baselines. The contributions can be highlighted as follows:
\begin{itemize}[leftmargin=*]
    \item \paratitle{New Perspective.} We study an understudied yet practical problem of source-free graph domain adaptation and propose a novel approach \method{} to solve the problem.
    \item \paratitle{New Method.} \method{} not only infers spectral seriation rankings to guide robust semantics learning under target domain label scarcity, but also detects harmonic graphs close to the source domain to guide subgraph learning for domain alignment and invariant learning. 
    \item \paratitle{Sate-of-the-art Performance.} Extensive experiments conducted on several benchmark datasets demonstrate the superiority of \method{} compared to extensive baselines.
\end{itemize}

\section{Related Work}

\paratitle{Graph neural networks~(GNNs)}~\cite{wu2020comprehensive,ju2024comprehensive} have made significant progress in a variety of tasks, including graph classification~\cite{coco}, visual grounding~\cite{transrefer3d,Luo_2022_CVPR}.
The majority of GNNs follow the message-passing mechanism~\cite{GIN}, which updates node information by aggregating from neighboring nodes. 
When applied to graph-level tasks, these updated node features are then combined into a comprehensive graph representation using various pooling techniques~\cite{lee2019self,bianchi2020spectral}, which can be utilized for downstream graph classification. However, these techniques~\cite{li2019semi} usually assume that training and test data are from the same distribution, which often does not hold in practical scenarios. 
To solve the out-of-distribution problem, some approaches have been proposed, where typically the source data is required~\cite{coco,icde2024,ju2024survey}. However, it is not practical to guarantee that the source data is available, due to, \eg, privacy and storage concerns.
To address this, our research focuses on source-free graph domain adaptation, which transfers the model to new domains without requiring access to the original data.

\paratitle{Source-free Domain Adaptation~(SFDA)}~\cite{fang2022source,yu2023comprehensive} eliminates the dependence on source data~\cite{yang2021exploiting,yang2021generalized,vs2023instance}. The assumption that adequate source data are available for adaptation does not always hold in real-world scenarios. On one hand, issues of privacy, confidentiality, and copyright restrictions may hinder access to the source data. On the other hand, storing the complete source dataset on devices with limited capacity is often impractical. Extensive approaches attempt to solve this problem, which can be classified into two primary categories, \ie, self-training approaches~\cite{sun2020test,saito2017asymmetric,shot} and generative approaches~\cite{nado2020evaluating,Schneider_Rusak_Eck_Bringmann_Brendel_Bethge_2020}. Self-training approaches usually adopt contrastive learning and pseudo-labeling for semantics learning.
Generative approaches typically reconstruct virtual samples using stored statistics in the source model and reduce the domain discrepancy. However, SFDA is under-explored for non-Euclidean graph data, where GALA~\cite{10561561} is the first attempt in the graph SFDA problem. In this paper, we detect harmonic graphs close to the source domain and extract subgraphs from inharmonic graphs for domain alignment. 

\section{The Proposed \method{}}

\begin{figure*}[!t]
    \centering
    \includegraphics[width=0.93\textwidth]{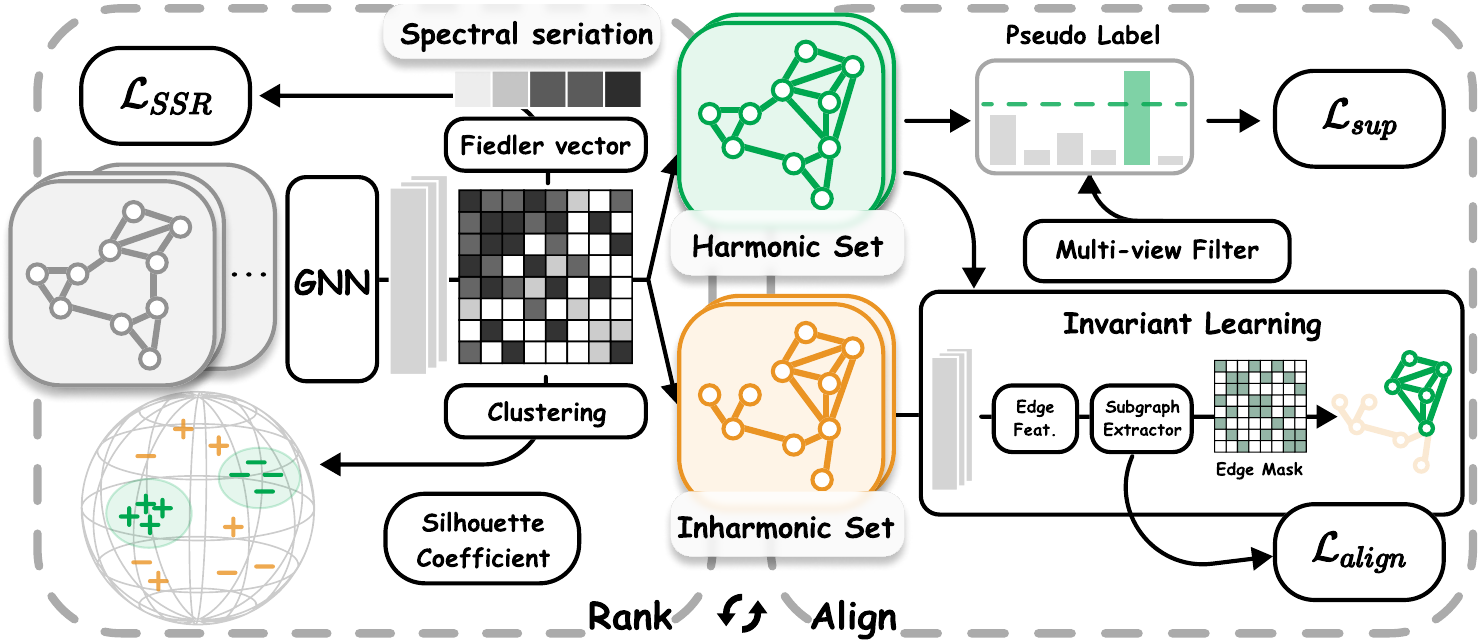}
    \vspace{-2mm}
    \caption{Overview of \method{}.
\textbf{\textit{(Left) Rank.}}~\method{} use the seriation similarity ranking learning for robust semantics learning under label scarcity, as in Section~\ref{sec:SSR}.
\textbf{\textit{(Right) Align.}}~\method{} detect the harmonic set and align the inharmonic set with a subgraph extractor and invariant learning, while applying discrimination learning with filtered pseudo-label, as in Section~\ref{sec:subgraph-align} and~\ref{sec:pseudo-label}. \textbf{\textit{(Center)} \method{}} alternates between two steps, achieving effective source-free domain adpation and addressing critical dilemmas.
    }
    \label{fig:fig2}
    \vspace{-2mm}
\end{figure*}

\subsection{Preliminaries}\label{sec:pre}

\paratitle{Problem Definition.}
A graph is represented as $G = (V, E)$, where $V$ is the set of nodes and $E$ is the set of edges. The nodes' attribute is denoted as $\bm{X} \in \mathbb{R}^{|V| \times d^f}$, where $|V|$ is the number of nodes, and $d^f$ is the dimension of a node's attribute. A dataset from the source domain is denoted as $\mathcal{D}^{so} = \{(G_i^{so}, y_i^{so})\}_{i=1}^{N_s}$, where $G_i^{so}$ is the $i$-th graph from the source and $y_i^{so}$ is its corresponding label. An unlabeled dataset from the target domain is represented as $\mathcal{D}^{ta} = \{G_j^{ta}\}_{j=1}^{N_t}$, where $G_j^{ta}$ is the $j$-th target sample. $y^{ta}_j$ is its corresponding label but is not available. Both domains share the same label space $\{1,\cdots, C\}$, but the data distributions are distinct. Our objective is to transfer a model pre-trained on the source domain to the target domain. Note that the source graphs are not available during the adaptation.

\paratitle{Graph Neural Networks.}
Given a node representation $\bm{h}_v^{(l)}$ for $v$ at layer $l$, the updating rule can be formalized as:
\begin{equation}
\boldsymbol{h}_{v}^{(l)}=\operatorname{COM}^{(l)}\left(\boldsymbol{h}_{v}^{(l-1)}, \operatorname{AGG}^{(l)}\left(\left\{\boldsymbol{h}_{u}^{(l-1)}\right\}_{u \in \mathcal{N}(v)}\right)\right) \,,
\end{equation}
in which $\operatorname{AGG}^{(l)}(\cdot)$ and $\operatorname{COM}^{(l)}(\cdot)$ represents the aggregation and combination operators respectively. $\mathcal{N}(v)$ collects all the neighbors connected with $v$. Finally, we introduce $\operatorname{READOUT}(\cdot)$ for node representation summarization:
\begin{equation}
\bm{z} = \operatorname{READOUT}\left(\left\{\boldsymbol{h}_{v}^{(K)}\right\}_{v \in V}\right) \,,
\end{equation}
where $K$ denotes the number of layers. A classifier $\operatorname{CLA}(\cdot)$ is adopted to map deep embedding into label distribution:
\begin{equation}
    \bm{p}^{(M)} = \operatorname{CLA} (\bm{z}) = \Phi_\theta(G) \,, 
\end{equation}
where $\Phi$ represents the entire graph neural network, pre-trained using the source graphs beforehand.

\subsection{Framework Overview}

This work studies the problem of Source-free Graph Domain Adaptation~(SFGDA), which is challenging due to target domain label scarcity and distribution shifts from source to target domain, without access to source graphs. We propose a novel approach named \method{} for this problem. In particular, to guide the semantics learning under label scarcity, \method{} introduces spectral seriation rankings among graph representations, which are robust to noise attack. In addition, \method{} utilizes spectral clustering and silhouette coefficients to identify harmonic graphs close to the source domain. Subgraphs with domain-invariant semantics are extracted in inharmonic graphs using an adversarial edge sampling process, which can guide invariant learning of GNNs. Pseudo-labeling on harmonic graphs with filtering is conducted to further enhance semantics learning. An overview of \method{} can be found in Figure~\ref{fig:fig2} and then we elaborate on the details. 

\subsection{Robust Semantics Learning with Seriation Similarity Rankings}\label{sec:SSR}

The significant challenge by SFGDA is label scarcity in the target domain. Due to serious distribution shifts, target data often contains noise and inaccurate pseudo-labels, leading to error accumulation and instability. \method{} addresses this issue from a spectral perspective, utilizing the intrinsic ranking information for better semantics learning. In this part, we will introduce our Seriation Similarity Rankings~(SSR) to guide semantics learning and theoretically demonstrate its robustness to noise.

We first extract the similarity matrix $S$ by,
\begin{equation}
\mS_{i, j} = \mathbf{sim}(\rvz_i, \rvz_j),
\end{equation}
where $\mathbf{sim}$ denotes the cosine similarity function, and $\rvz_i$ and $\rvz_j$ represent the graph representations as Section~\ref{sec:pre}. Then, we employ the spectral seriation method~\cite{atkins1998spectral} to obtain the data ranking matrix $\mathbf{R}$ within unlabeled target data, from the correlation matrix. The higher correlation values suggest a closer proximity in ranking.
Spectral seriation can be effectively utilized to reconstruct $\mR$ from $\mS$, since the cosine similarity corresponds to correlation following $L_2$ normalization.
The seriation ranking can be derived by minimizing the following equation:
\begin{equation}
\argmin\limits_\mR \sum_{i,j} \mS_{i,j}(\mR_i-\mR_j)^2,
\end{equation}
which brings sample pairs with a higher degree of correlation closer to the learned representation space, by minimizing the loss function that encourages proximity between $\mR_i$ and $\mR_j$. 
According to the spectral seriation~\cite{atkins1998spectral,gong2022ranksim,dai2024semi}, the optimal ranking $\mR$ can be obtained from the {Fiedler vector}. This is formalized in Theorem~\ref{theorem:spectral}.
\begin{theorem} \label{theorem:spectral} % [Spectral Seriation]
The seriation ranking that most accurately reflects observed $\mS$ is the ranking of the values in the Fiedler vector $\lambda$ of the Laplacian matrix $\mL$ by$$\mR = sorting(\lambda).$$
\end{theorem}
The Fiedler vector $\lambda$ is the eigenvector corresponding to the second smallest eigenvalue of the Laplacian matrix $\mL = \mathbf{diag}(\mS\mathbf{1}) - \mS$.
The ranking $\mR$ given from the spectral seriation can be used to guide semantics learning by
\begin{equation}\label{eq:loss-ssr}
    \mathcal{L}_{SSR} = \sum^{\lvert B\rvert}_{i=1} \mathbf{rsim}(\mathbf{rk}(\mS_{[i,:]}), \mathbf{rk}(-\lvert \mR-\mR[i]\rvert)),
\end{equation}
where $B$ is the sampled batch, $[i]$ denotes the $i$-th value of a vector, $\mathbf{rk}$ denotes the ranking operator. $\mathbf{rsim}$ is a differentiable ranking similarity function which is the differential combinatorial solver~\cite{poganvcic2019differentiation}.

\paratitle{Robustness under Perturbation.} In this part, we will discuss the inherent noise-robustness of the spectral seriation method theoretically by deriving the approximate bounds. 

We first give the perturbation bounds for eigenvalues and the upper bound for the Fiedler value. 
Let $\mA, \mB \in \sC$ be Hermitian matrices. Let the eigenvalues of $\mA$ and $\mB$ are ${\lambda_i}$ and $ {\mu_i}$, with the eigenvalues sorted in non-decreasing order. 
Then, for each corresponding pair of eigenvalues, the following inequality holds on the perturbation bounds for the eigenvalues:
\begin{equation}
\lvert \mu_i - \lambda_i \rvert \leq \lVert \mB - \mA \rVert_2.
\end{equation}
Moreover, when $\lambda$ is the Fiedler value of the {Laplacian} matrix $\rmL$ from the similarity matrix $\mS$, then the upper bound of the $\lambda$ is given by
\begin{equation}
\lambda \leq \frac{n}{n-1} \min_{1\leq i\leq n} \{\rmL_{ii}\}.
\end{equation}
Then we analyze the potential noise on the similarity matrix. 
\begin{theorem}
Consider the perturbation matrix of $\mS$ is $\Delta \mS$, when
$
2\lVert \Delta \mS \rVert_F \leq 1 - \frac{\min_{1\leq i\leq n} {\sum_{t\neq i} {\lvert \mS_{it} \rvert}}}{n-1},
$
the ranking obtained by the SSR algorithm using $\mS$ is the same as that obtained by the SSR algorithm using $\mS+\Delta \mS.$
\end{theorem}
$\lVert \cdot \rVert_F$ denotes the {Frobenius} norm. The theorem utilizes the first-order approximation to establish a bound that the spectral ranking can tolerate without altering results. 

In summary, given the significant noise in the target domain, SSR provides theoretically error-bounded data relationships. This allows us to obtain better target domain representations through similarity learning.

\subsection{Subgraph Extraction for Domain Alignment}\label{sec:subgraph-align}

To address label scarcity of target graphs, \method{} detects harmonic graphs that are close to the source domain from a spectral perspective. Then, we extract subgraphs in inharmonic graphs with an adversarial edge sampling process, which guides effective invariant learning for GNNs. 

\paratitle{Harmonic Graph Detection.}
\method{} uses spectral clustering and silhouette coefficients to discover the harmonic graphs with the source domain. This process is instrumental in guiding the domain alignment for subgraph extraction that has domain-invariant semantics.

In detail, we utilize the similarity matrix $\mS$ and the \textit{Laplacian} matrix $\mL$ computed in Section~\ref{sec:SSR} to obtain the eigenvectors $\mU$. To compute the eigenvectors, we address the generalized eigenvalue problem:
\begin{equation}
\mL \mU = \mU \mathbf{\Lambda}\,,
\end{equation}
where $\mU \in \mathbb{R}^{n \times k}$ represents the matrix of eigenvectors corresponding to the $k$ smallest eigenvalues, and $\mathbf{\Lambda} \in \mathbb{R}^{k \times k}$ is the diagonal matrix containing these eigenvalues. We then apply {k-means} clustering algorithm to the rows of $\mU$, to obtain the clustering labels $\rvc$. After the clustering, we calculate the silhouette coefficient to detect the harmonic graphs, which are close to the source graphs. For a data point $i$ within a cluster, the silhouette coefficient $S(i)$~\cite{silhouettes,monshizadeh2022deep} is :
\begin{equation} \label{eq:sc}
S(i) = \frac{b(i) - a(i)}{\max\{a(i), b(i)\}}\,,
\end{equation}
where $a(i)$ is the mean dissimilarity of $i$ to all other points in the same cluster, while $b(i)$ is the smallest mean dissimilarity to all points in any other cluster. Since $S(i) \in [-1, 1]$, a coefficient close to $1$ indicates that the data point is well-matched to its cluster and distinctly separated from neighboring clusters. These graphs have the potential to be well-classified using source GNNs, which are more close to the source domain. Therefore, we rank the target graphs by silhouette coefficients and select the top $\rho$ graphs to construct a harmonic set. $\rho$ is set heuristically to $40\%$.

\paratitle{Domain-invariant Subgraph Extraction.}
After partitioning the dataset, we utilize the harmonic set to simulate the unavailable source graphs to depict domain discrepancy. To identify crucial parts invariant to different domains, we sample edges for inharmonic graphs, which extract domain-invariant subgraphs using adversarial learning. Moreover, invariant learning is introduced to encourage consistent predictions of GNNs after removing redundant information. 

In particular, we use a neural subgraph extractor $f_\theta$. For edge embedding $\rve$ concatenated from node embedding $\rvh$,
we predict the sampled adjacency matrix $\hat{\mA}_{sub}$ with:
\begin{equation}
\hat{\mA}_{sub} = f_\theta(\rve, \mA)\,,
\end{equation}
where $\mA$ is the original adjacency matrix.
Since the sampling process is not differentiable, we apply Gumbel-Sigmoid~\cite{gumbel} for differentiable training. Then, we extract the subgraph $G_{sub}$ according to $\hat{\mA}_{sub}$. 

To train the subgraph discovery network, we adopt an adversarial learning framework. The discriminator is to distinguish the domain of the input graph, while the subgraph discovery network is used as the adversarial component, and aims to extract domain-invariant subgraphs that confuse the discriminator. Let $D$ denote the discriminator and $G$ denote the subgraph discovery network. The training loss can be formulated as follows:
\begin{equation}\label{eq:loss-adv}
\begin{aligned}
\mathcal{L}_{\text{adv}}(f_\theta, D) = \mathbb{E}_{G_h \sim p_{\text{data}}(G_h)}[\log D(G_h)] + \\
\mathbb{E}_{G_i \sim p_{\text{data}}(G_i)}[\log(1 - D(f_\theta(G_i)))]\,,
\end{aligned}
\end{equation}
where $G_h$ and $G_i$ are the harmonic and inharmonic graphs respectively, $p_{\text{data}}(\cdot)$ represent the data distributions. 
Subgraph extractor can effectively extract domain-invariant components under domain shift. Then, we introduce invariant learning to encourage the consistency of predictions after extracting domain-invariant subgraphs, which ensures GNNs are insensitive to the domain-specific information on the target domain. In formulation, the invariant learning objective is defined as:
\begin{equation}\label{eq:loss-inv}
\mathcal{L}_{\text{inv}} = - \frac{1}{|\sI|} \sum_{G_i^{ta} \in \sI} \text{KL}(\tilde{\bm{p}}_i^{ta}\parallel{\bm{p}}_i^{ta})\,,
\end{equation}
where $\sI$ is the inharmonic set, $\tilde{\bm{p}}_i^{ta}$ is the label distribution of the subgraph $f_\theta(G_i)$. When the KL divergence is minimized, \method{} enables the model to leverage the inharmonic graphs, which are initially more challenging due to the domain shift, and gradually adapt to the target domain.

Finally, the overall training objective for the subgraph discovery network can be formulated as a combination of the adversarial loss and the invariant learning loss:
\begin{equation}\label{eq:loss-align}
\mathcal{L}_{\text{align}} = \mathcal{L}_{\text{adv}}+ \mathcal{L}_{\text{inv}}\,.
\end{equation}
By jointly optimizing the subgraph discovery network and the discriminator, we can effectively extract domain-invariant subgraphs that confuse the discriminator while preserving the original semantics.

\subsection{Semantics Enhancement with Filtered Pseudo-labeling}\label{sec:pseudo-label}

To enhance the semantics learning under label scarcity and potential noise, we propose to obtain the confident discriminative learning set from multi-view filtering. 
From the global view, we have detected harmonic graphs on the target domain. Then, from the local view, we leverage the predicted confidence scores to select reliable samples. The combination of these two perspectives yields a more confident set.

To filter samples in the harmonic set $\sH$ and generate the confident set $\sC$, we define a threshold $\tau$ for the confidence scores. Samples with confidence scores exceeding $\tau$ are considered to be part of the confidence set. Formally, the confident set $\sC$ is defined as:
\begin{equation}\label{eq:confident-set}
    \sC = \{ G_j^{ta} \in \sH \mid \max(\bm{p}_j^{ta}) \geq \tau \},
\end{equation}
where $\bm{p}_j^{ta}$ represents the model prediction, indicating the predicted probability distribution over classes, and $\tau$ is the pre-defined threshold. 

Then, the standard cross-entropy loss is minimized in the confident set $\sC$:
\begin{equation}\label{eq:loss-sup}
    \mathcal{L}_{sup} = - \frac{1}{\lvert\sH\rvert} \sum_{G_j^{ta} \in \sC} \log \bm{p}_j^{ta}[\hat{y}_j^{ta}],
\end{equation}
where $\hat{y}_j^{ta}$ denotes the pseudo-label of $G_j^{ta}$. Our pseudo-labeling strategy provides a reliable optimization process with reduced error accumulation. After a certain adaptation period, we re-perform the harmonic set partitioning operation to include more confident data into the harmonic set. This iterative process helps to gradually enhance semantics learning by incorporating more reliable pseudo-labeled data.

As a preliminary, we adopt an off-the-shelf model trained on the source domain. Then, the loss objectives are minimized in the target data, which is summarized in Algorithm~\ref{alg1}.

\paratitle{Complexity Analysis.}
The computational complexity of seriation similarity and spectral clustering is $\mathcal{O}(|D^{ta}|^2d)$, where $|D^{ta}|$ is the target graph amount, $d$ is the feature dimension. Given a graph $G = (V, E)$, $\lVert A\rVert_0$ is the number of nonzero entries in the adjacency matrix, $|V|$ is the number of nodes, and $K$ is the number of GNN layers. The complexity of the GNN is $\mathcal{O}(K|V|{d}^2)$. The subgraph extractor takes $\mathcal{O}(K\lVert A\rVert_0{d})$. Therefore, the complexity for each sample is $\mathcal{O}(K|V|{d}^2 + K\lVert A\rVert_0{d})$, which is linear to $\lVert A\rVert_0$ and $|V|$. 

\begin{algorithm}[tb]
    \caption{Optimization Algorithm of \method{}}\label{alg1}
    \label{alg:algorithm}
    \textbf{Input}: Pre-trained model $\Phi(\cdot)$, target graphs ${\gD}^{ta}$, \\
    \textbf{Output}: GNN-based model $\Phi(\cdot)$

    \begin{algorithmic}[1] 
    \FOR{epoch = 1, 2, $\cdots$}
    
    \STATE Generate harmonic and inharmonic sets using Eq.~\ref{eq:sc};
    \FOR{each batch}
    \STATE Sample a mini-batch from ${\gD}^{ta}$;
    \STATE Calculate the loss objective using Eq.~\ref{eq:loss-ssr};
    \STATE \textit{// For the harmonic set}\\
    \STATE Calculate the discriminating loss using Eq.~\ref{eq:loss-sup};
    \STATE \textit{// For the inharmonic set}\\
    \STATE Calculate the discriminating loss using Eq.~\ref{eq:loss-align};
    \STATE Back-propagation;
    \ENDFOR
    \ENDFOR
    \end{algorithmic}
    % \vspace{-2mm}
\end{algorithm}

\begin{table*}[t]
\centering
\tabcolsep=2pt
% \vspace{-1mm}
\resizebox{\textwidth}{!}{
\begin{tabular}{llllllllllllll}
\toprule[1pt]
{\bf Methods} &M0$\rightarrow$M1 &M1$\rightarrow$M0 &M0$\rightarrow$M2 &M2$\rightarrow$M0 &M0$\rightarrow$M3 &M3$\rightarrow$M0 &M1$\rightarrow$M2 &M2$\rightarrow$M1 &M1$\rightarrow$M3 &M3$\rightarrow$M1 &M2$\rightarrow$M3 &M3$\rightarrow$M2 & \textbf{Avg.}\\
\midrule
\midrule
GCN &
% 0-1
68.0{\tiny $\pm 2.0$}
&
% 1-0
68.8{\tiny $\pm 1.5$}
&
% 0-2
60.5{\tiny $\pm 2.8$}
&
% 2-0
64.4{\tiny $\pm 1.5$}
&
% 0-3
53.7{\tiny $\pm 1.3$}
&
% 3-0
58.1{\tiny $\pm 1.4$}
&
% 1-2
75.2{\tiny $\pm 0.8$}
&
% 2-1
76.2{\tiny $\pm 1.5$}
&
% 1-3
\textbf{67.5}{\tiny $\pm 1.2$}
&
% 3-1
55.4{\tiny $\pm 1.5$}
&
% 2-3
62.5{\tiny $\pm 1.0$}
&
% 3-2
\textbf{68.5}{\tiny $\pm 1.5$}
&
% avg
64.9{\tiny $\pm 1.5$}
\\
GIN &
% 0-1
70.6{\tiny $\pm 0.4$}
&
% 1-0
64.2{\tiny $\pm 0.9$}
&
% 0-2
63.5{\tiny $\pm 1.2$}
&
% 2-0
62.5{\tiny $\pm 0.7$}
&
% 0-3
57.0{\tiny $\pm 0.1$}
&
% 3-0
56.4{\tiny $\pm 0.3$}
&
% 1-2
73.3{\tiny $\pm 0.5$}
&
% 2-1
76.5{\tiny $\pm 0.4$}
&
% 1-3
65.2{\tiny $\pm 0.6$}
&
% 3-1
53.3{\tiny $\pm 1.8$}
&
% 2-3
64.4{\tiny $\pm 0.8$}
&
% 3-2
66.8{\tiny $\pm 0.5$}
&
% avg
64.5{\tiny $\pm 0.7$}
\\
GraphSAGE &
% 0-1
71.2{\tiny $\pm 0.6$}
&
% 1-0
65.6{\tiny $\pm 0.3$}
&
% 0-2
64.3{\tiny $\pm 0.3$}
&
% 2-0
65.5{\tiny $\pm 0.2$}
&
% 0-3
57.3{\tiny $\pm 0.5$}
&
% 3-0
56.5{\tiny $\pm 0.3$}
&
% 1-2
74.7{\tiny $\pm 0.4$}
&
% 2-1
77.6{\tiny $\pm 0.7$}
&
% 1-3
62.3{\tiny $\pm 0.4$}
&
% 3-1
51.7{\tiny $\pm 0.3$}
&
% 2-3
62.4{\tiny $\pm 0.5$}
&
% 3-2
62.3{\tiny $\pm 0.7$}
&
% avg
64.3{\tiny $\pm 0.4$}
\\
GAT &
% 0-1
69.7{\tiny $\pm 1.0$}
&
% 1-0
67.0{\tiny $\pm 1.6$}
&
% 0-2
62.7{\tiny $\pm 2.3$}
&
% 2-0
67.0{\tiny $\pm 1.5$}
&
% 0-3
56.1{\tiny $\pm 2.1$}
&
% 3-0
57.8{\tiny $\pm 1.7$}
&
% 1-2
\textbf{76.6}{\tiny $\pm 1.2$}
&
% 2-1
77.2{\tiny $\pm 0.4$}
&
% 1-3
63.4{\tiny $\pm 1.3$}
&
% 3-1
53.0{\tiny $\pm 3.9$}
&
% 2-3
60.7{\tiny $\pm 0.6$}
&
% 3-2
61.8{\tiny $\pm 3.2$}
&
% avg
64.6{\tiny $\pm 1.7$}
\\
Mean-Teacher &
% 0-1
65.3{\tiny $\pm 4.7$}
&
% 1-0
52.1{\tiny $\pm 3.4$}
&
% 0-2
\textbf{70.6}{\tiny $\pm 2.5$}
&
% 2-0
52.2{\tiny $\pm 1.9$}
&
% 0-3
49.9{\tiny $\pm 0.5$}
&
% 3-0
49.0{\tiny $\pm 0.4$}
&
% 1-2
66.2{\tiny $\pm 1.4$}
&
% 2-1
62.7{\tiny $\pm 4.1$}
&
% 1-3
50.1{\tiny $\pm 1.3$}
&
% 3-1
72.2{\tiny $\pm 1.6$}
&
% 2-3
48.9{\tiny $\pm 1.7$}
&
% 3-2
65.8{\tiny $\pm 3.1$}
&
% avg
58.7{\tiny $\pm 2.2$}
\\
InfoGraph &
% 0-1
69.1{\tiny $\pm 1.8$}
&
% 1-0
\textbf{68.9}{\tiny $\pm 0.3$}
&
% 0-2
66.6{\tiny $\pm 2.5$}
&
% 2-0
64.9{\tiny $\pm 1.2$}
&
% 0-3
55.9{\tiny $\pm 1.0$}
&
% 3-0
57.8{\tiny $\pm 2.1$}
&
% 1-2
74.7{\tiny $\pm 0.3$}
&
% 2-1
76.8{\tiny $\pm 1.5$}
&
% 1-3
65.6{\tiny $\pm 0.6$}
&
% 3-1
57.1{\tiny $\pm 3.2$}
&
% 2-3
64.7{\tiny $\pm 1.9$}
&
% 3-2
64.2{\tiny $\pm 2.9$}
&
% avg
65.5{\tiny $\pm 1.6$}
\\
TGNN &
% 0-1
73.3{\tiny $\pm 4.9$}
&
% 1-0
61.9{\tiny $\pm 2.4$}
&
% 0-2
65.3{\tiny $\pm 4.0$}
&
% 2-0
58.1{\tiny $\pm 2.4$}
&
% 0-3
55.5{\tiny $\pm 3.5$}
&
% 3-0
58.1{\tiny $\pm 2.4$}
&
% 1-2
65.9{\tiny $\pm 1.1$}
&
% 2-1
66.7{\tiny $\pm 3.9$}
&
% 1-3
66.5{\tiny $\pm 2.1$}
&
% 3-1
70.1{\tiny $\pm 1.0$}
&
% 2-3
55.5{\tiny $\pm 3.5$}
&
% 3-2
65.3{\tiny $\pm 3.0$}
&
% avg
63.5{\tiny $\pm 2.9$}
\\
PLUE &
% 0-1
75.2{\tiny $\pm 1.4$}
&
% 1-0
68.5{\tiny $\pm 0.5$}
&
% 0-2
66.3{\tiny $\pm 1.1$}
&
% 2-0
67.9{\tiny $\pm 1.6$}
&
% 0-3
54.0{\tiny $\pm 1.3$}
&
% 3-0
56.4{\tiny $\pm 1.4$}
&
% 1-2
68.4{\tiny $\pm 1.0$}
&
% 2-1
76.9{\tiny $\pm 3.5$}
&
% 1-3
62.9{\tiny $\pm 0.4$}
&
% 3-1
57.6{\tiny $\pm 3.0$}
&
% 2-3
62.0{\tiny $\pm 0.6$}
&
% 3-2
67.4{\tiny $\pm 2.6$}
&
% avg
65.3{\tiny $\pm 1.5$}
\\
\midrule
\rowcolor{LightCyan} \textbf{\method{}} &
% 0-1
\textbf{81.0}{\tiny $\pm 1.5$}
&
% 1-0
68.3{\tiny $\pm 2.6$}
&
% 0-2
68.4{\tiny $\pm 2.7$}
&
% 2-0
\textbf{68.8}{\tiny $\pm 1.3$}
&
% 0-3
\textbf{63.1}{\tiny $\pm 2.9$}
&
% 3-0
\textbf{59.8}{\tiny $\pm 2.1$}
&
% 1-2
72.7{\tiny $\pm 2.4$}
&
% 2-1
\textbf{80.9}{\tiny $\pm 1.7$}
&
% 1-3
62.1{\tiny $\pm 1.9$}
&
% 3-1
\textbf{73.4}{\tiny $\pm 2.9$}
% 73.4{\tiny $\pm 2.9$}
&
% 2-3
\textbf{65.1}{\tiny $\pm 2.4$}
&
% 3-2
67.8{\tiny $\pm 2.3$}
&
% avg
\textbf{69.3}{\tiny $\pm 2.2$}
\\
\bottomrule[1pt]
\end{tabular}
}
\vspace{-1mm}
\caption{The classification accuracy~(\%) on Mutagenicity (source$\rightarrow$target), where M0, M1, M2, and M3 are the sub-datasets.}
\vspace{-2mm}
\label{tab::muta}
\end{table*}

\begin{table*}[t]
\centering
\tabcolsep=2pt
% \vspace{-1mm}
\resizebox{\textwidth}{!}{
\begin{tabular}{llllllllllllll}
\toprule[1pt]
{\bf Methods} &P0$\rightarrow$P1 &P1$\rightarrow$P0 &P0$\rightarrow$P2 &P2$\rightarrow$P0 &P0$\rightarrow$P3 &P3$\rightarrow$P0 &P1$\rightarrow$P2 &P2$\rightarrow$P1 &P1$\rightarrow$P3 &P3$\rightarrow$P1 &P2$\rightarrow$P3 &P3$\rightarrow$P2 &\textbf{Avg.}\\
\midrule
\midrule
GCN &
% 0-1
71.9{\tiny $\pm 0.9$}
&
% 1-0
74.7{\tiny $\pm 2.9$}
&
% 0-2
62.6{\tiny $\pm 1.2$}
&
% 2-0
68.3{\tiny $\pm 3.8$}
&
% 0-3
51.1{\tiny $\pm 3.3$}
&
% 3-0
45.8{\tiny $\pm 3.1$}
&
% 1-2
57.6{\tiny $\pm 2.1$}
&
% 2-1
70.4{\tiny $\pm 2.2$}
&
% 1-3
39.7{\tiny $\pm 3.6$}
&
% 3-1
49.7{\tiny $\pm 2.6$}
&
% 2-3
58.3{\tiny $\pm 1.3$}
&
% 3-2
52.9{\tiny $\pm 3.0$}
&
% avg
58.6{\tiny $\pm 2.5$}
\\
GIN &
% 0-1
70.0{\tiny $\pm 2.1$}
&
% 1-0
60.7{\tiny $\pm 3.6$}
&
% 0-2
61.8{\tiny $\pm 2.6$}
&
% 2-0
72.9{\tiny $\pm 2.7$}
&
% 0-3
47.1{\tiny $\pm 3.3$}
&
% 3-0
44.3{\tiny $\pm 4.2$}
&
% 1-2
62.5{\tiny $\pm 2.1$}
&
% 2-1
68.9{\tiny $\pm 2.0$}
&
% 1-3
41.1{\tiny $\pm 3.2$}
&
% 3-1
47.9{\tiny $\pm 3.3$}
&
% 2-3
48.6{\tiny $\pm 4.0$}
&
% 3-2
56.1{\tiny $\pm 2.6$}
&
% avg
57.4{\tiny $\pm 3.0$}
\\
GraphSAGE &
% 0-1
72.2{\tiny $\pm 1.0$}
&
% 1-0
78.3{\tiny $\pm 3.0$}
&
% 0-2
64.7{\tiny $\pm 2.3$}
&
% 2-0
67.2{\tiny $\pm 1.1$}
&
% 0-3
46.9{\tiny $\pm 1.1$}
&
% 3-0
42.2{\tiny $\pm 4.0$}
&
% 1-2
62.6{\tiny $\pm 1.8$}
&
% 2-1
69.7{\tiny $\pm 0.8$}
&
% 1-3
32.9{\tiny $\pm 2.0$}
&
% 3-1
50.8{\tiny $\pm 2.3$}
&
% 2-3
56.1{\tiny $\pm 3.5$}
&
% 3-2
56.9{\tiny $\pm 3.4$}
&
% avg
58.9{\tiny $\pm 2.2$}
\\
GAT &
% 0-1
70.0{\tiny $\pm 3.7$}
&
% 1-0
71.4{\tiny $\pm 3.7$}
&
% 0-2
66.8{\tiny $\pm 2.0$}
&
% 2-0
73.9{\tiny $\pm 2.6$}
&
% 0-3
49.3{\tiny $\pm 1.3$}
&
% 3-0
40.4{\tiny $\pm 2.8$}
&
% 1-2
61.4{\tiny $\pm 4.6$}
&
% 2-1
68.9{\tiny $\pm 2.2$}
&
% 1-3
44.3{\tiny $\pm 3.8$}
&
% 3-1
49.3{\tiny $\pm 3.6$}
&
% 2-3
50.7{\tiny $\pm 2.4$}
&
% 3-2
52.1{\tiny $\pm 2.9$}
&
% avg
58.3{\tiny $\pm 3.0$}
\\
Mean-Teacher &
% 0-1
64.3{\tiny $\pm 4.1$}
&
% 1-0
71.4{\tiny $\pm 5.2$}
&
% 0-2
60.4{\tiny $\pm 3.6$}
&
% 2-0
72.1{\tiny $\pm 4.4$}
&
% 0-3
25.0{\tiny $\pm 5.6$}
&
% 3-0
55.4{\tiny $\pm 4.0$}
&
% 1-2
61.1{\tiny $\pm 2.3$}
&
% 2-1
60.7{\tiny $\pm 5.3$}
&
% 1-3
29.6{\tiny $\pm 4.8$}
&
% 3-1
49.3{\tiny $\pm 3.1$}
&
% 2-3
31.8{\tiny $\pm 4.0$}
&
% 3-2
55.4{\tiny $\pm 4.9$}
&
% avg
51.0{\tiny $\pm 4.3$}
\\
InfoGraph &
% 0-1
\textbf{74.0}{\tiny $\pm 2.7$}
&
% 1-0
77.6{\tiny $\pm 2.9$}
&
% 0-2
68.3{\tiny $\pm 3.6$}
&
% 2-0
71.1{\tiny $\pm 1.1$}
&
% 0-3
46.9{\tiny $\pm 3.2$}
&
% 3-0
46.5{\tiny $\pm 2.1$}
&
% 1-2
64.4{\tiny $\pm 1.5$}
&
% 2-1
\textbf{72.2}{\tiny $\pm 1.9$}
&
% 1-3
41.9{\tiny $\pm 1.1$}
&
% 3-1
35.4{\tiny $\pm 1.9$}
&
% 2-3
54.7{\tiny $\pm 3.1$}
&
% 3-2
\textbf{62.6}{\tiny $\pm 4.2$}
&
% avg
59.4{\tiny $\pm 2.4$}
\\
TGNN &
% 0-1
60.4{\tiny $\pm 3.4$}
&
% 1-0
41.8{\tiny $\pm 2.8$}
&
% 0-2
45.7{\tiny $\pm 4.6$}
&
% 2-0
46.6{\tiny $\pm 4.2$}
&
% 0-3
\textbf{64.8}{\tiny $\pm 5.4$}
&
% 3-0
42.0{\tiny $\pm 3.9$}
&
% 1-2
58.6{\tiny $\pm 3.2$}
&
% 2-1
58.4{\tiny $\pm 3.7$}
&
% 1-3
\textbf{75.7}{\tiny $\pm 4.0$}
&
% 3-1
\textbf{68.4}{\tiny $\pm 3.2$}
&
% 2-3
\textbf{65.5}{\tiny $\pm 4.8$}
&
% 3-2
53.8{\tiny $\pm 2.0$}
&
% avg
55.9{\tiny $\pm 3.8$}
\\
PLUE &
% 0-1
64.3{\tiny $\pm 2.3$}
&
% 1-0
71.5{\tiny $\pm 2.1$}
&
% 0-2
65.0{\tiny $\pm 3.0$}
&
% 2-0
78.4{\tiny $\pm 2.1$}
&
% 0-3
45.9{\tiny $\pm 3.7$}
&
% 3-0
\textbf{63.7}{\tiny $\pm 3.0$}
&
% 1-2
58.3{\tiny $\pm 3.3$}
&
% 2-1
68.9{\tiny $\pm 2.7$}
&
% 1-3
41.9{\tiny $\pm 1.3$}
&
% 3-1
45.7{\tiny $\pm 4.1$}
&
% 2-3
47.5{\tiny $\pm 3.2$}
&
% 3-2
58.3{\tiny $\pm 2.7$}
&
% avg
59.1{\tiny $\pm 2.8$}
\\
\midrule 
% 92FA7A, D6FDD0
\rowcolor{LightCyan} \textbf{\method{}} &
% 0-1
69.7{\tiny $\pm 1.8$}
&
% 1-0
\textbf{79.6}{\tiny $\pm 1.3$}
&
% 0-2
\textbf{72.5}{\tiny $\pm 2.6$}
&
% 2-0
\textbf{78.5}{\tiny $\pm 1.8$}
&
% 0-3
64.4{\tiny $\pm 3.6$}
&
% 3-0
57.8{\tiny $\pm 5.5$}
&
% 1-2
\textbf{68.9}{\tiny $\pm 3.3$}
&
% 2-1
67.8{\tiny $\pm 3.5$}
&
% 1-3
56.5{\tiny $\pm 3.8$}
&
% 3-1
57.8{\tiny $\pm 5.3$}
&
% 2-3
65.3{\tiny $\pm 3.3$}
&
% 3-2
55.6{\tiny $\pm 3.4$}
&
% avg
\textbf{66.2}{\tiny $\pm 3.2$}
\\
\bottomrule[1pt]
\end{tabular}}
\vspace{-1mm}
\caption{The classification accuracy~(\%) on PROTEINS (source$\rightarrow$target), where P0, P1, P2, and P3 are the sub-datasets.}
\vspace{-2mm}
\label{tab::protein}
\end{table*}

\begin{table}[t]
\centering
\tabcolsep=2pt
\resizebox{\linewidth}{!}{
\begin{tabular}{llllll}
\toprule[1pt]
{\bf Methods} &C$\rightarrow$CM & CM$\rightarrow$C &B$\rightarrow$BM &BM$\rightarrow$B &\textbf{Avg.}\\
\midrule
\midrule
GCN &
% 0-1
54.1{\tiny $\pm 2.6$}&
% 1-0
46.6 {\tiny $\pm 4.1$}&
% 0-2
51.3 {\tiny $\pm 2.3$}&
% 2-0
62.8 {\tiny $\pm 3.7$}&
% avg
53.7{\tiny $\pm 3.2$}
\\
GIN &
% 0-1
51.1{\tiny $\pm 2.2$}&
% 1-0
46.4{\tiny $\pm 4.6$}&
% 0-2
48.1{\tiny $\pm 3.6$}&
% 2-0
65.6{\tiny $\pm 2.8$}&
% avg
52.8{\tiny $\pm 3.3$}
\\
GraphSAGE &
% 0-1
49.2{\tiny $\pm 3.4$}&
% 1-0
42.9{\tiny $\pm 3.9$}&
% 0-2
47.3{\tiny $\pm 1.5$}&
% 2-0
67.5{\tiny $\pm 3.1$}&
% avg
51.7{\tiny $\pm 3.0$}
\\
GAT &
% 0-1
52.0{\tiny $\pm 1.8$}&
% 1-0
48.9{\tiny $\pm 3.7$}&
% 0-2
48.4{\tiny $\pm 2.2$}&
% 2-0
61.3{\tiny $\pm 4.2$}&
% avg
52.6{\tiny $\pm 3.0$}
\\
% \midrule
Mean-Teacher &
% 0-1
53.0{\tiny $\pm 2.3$}&
% 1-0
42.6{\tiny $\pm 4.9$}&
% 0-2
50.6{\tiny $\pm 2.1$}&
% 2-0
57.6{\tiny $\pm 4.3$}&
% avg
50.9{\tiny $\pm 3.4$}
\\
InfoGraph &
% 0-1
45.9{\tiny $\pm 3.4$}&
% 1-0
48.9{\tiny $\pm 3.3$}&
% 0-2
51.9{\tiny $\pm 3.2$}&
% 2-0
65.2{\tiny $\pm 4.7$}&
% avg
53.4{\tiny $\pm 3.6$}
\\
TGNN &
% 0-1
48.3{\tiny $\pm 4.2$}&
% 1-0
52.1{\tiny $\pm 5.6$}&
% 0-2
46.4{\tiny $\pm 2.7$}&
% 2-0
68.5{\tiny $\pm 4.3$}&
% avg
53.8{\tiny $\pm 4.2$}
\\
% \midrule
PLUE &
% 0-1
54.4{\tiny $\pm 1.4$}&
% 1-0
41.7{\tiny $\pm 3.2$}&
% 0-2
49.8{\tiny $\pm 2.8$}&
% 2-0
\textbf{74.8}{\tiny $\pm 1.5$}&
% avg
55.2{\tiny $\pm 2.2$}
\\
\midrule
\rowcolor{LightCyan}
{\bf \method{}} &
% 0-1
\textbf{57.2}{\tiny $\pm 2.8$}
&
% 1-0
\textbf{67.8}{\tiny $\pm 3.3$}
&
% 0-2
\textbf{52.3}{\tiny $\pm 2.6$}
&
% 2-0
73.5{\tiny $\pm 3.0$}
&
% avg
\textbf{62.7}{\tiny $\pm 2.9$}
\\
\bottomrule[1pt]
\end{tabular}
}
\vspace{-1mm}
\caption{The classification accuracy~(\%) on COX2 and BZR~(source$\rightarrow$target). C, CM, B, and BM are for COX2, COX2\_MD, BZR, and BZR\_MD datasets.}
\vspace{-2mm}
\label{tab::cross}
\end{table}

\begin{table*}[t]
\centering
\tabcolsep=2pt
\resizebox{\textwidth}{!}{
\begin{tabular}{llllllllllllll}
\toprule[1pt]
{\bf Methods} &F0$\rightarrow$F1 &F1$\rightarrow$F0 &F0$\rightarrow$F2 &F2$\rightarrow$F0 &F0$\rightarrow$F3 &F3$\rightarrow$F0 &F1$\rightarrow$F2 &F2$\rightarrow$F1 &F1$\rightarrow$F3 &F3$\rightarrow$F1 &F2$\rightarrow$F3 &F3$\rightarrow$F2 &\textbf{Avg.}\\
\midrule
\midrule
GCN &
% 0-1
55.3{\tiny $\pm 0.8$} 
&
% 1-0
\textbf{56.4}{\tiny $\pm 1.6$} 
&
% 0-2
60.4{\tiny $\pm 1.9$}
&
% 2-0
54.6{\tiny $\pm 1.4$}
&
% 0-3
46.7{\tiny $\pm 1.8$} 
&
% 3-0
51.6{\tiny $\pm 1.7$}
&
% 1-2
60.7{\tiny $\pm 0.8$}
&
% 2-1
58.3{\tiny $\pm 1.3$} 
&
% 1-3
47.9{\tiny $\pm 1.8$}
&
% 3-1
47.5{\tiny $\pm 0.8$}
&
% 2-3
52.1{\tiny $\pm 2.6$} 
&
% 3-2
54.6{\tiny $\pm 1.8$} 
&
% avg
53.8{\tiny $\pm 1.5$} 
\\
GIN &
% 0-1
56.5{\tiny $\pm 0.5$}
&
% 1-0
53.8{\tiny $\pm 1.4$}
&
% 0-2
57.2{\tiny $\pm 0.4$}
&
% 2-0
57.9{\tiny $\pm 1.8$}
&
% 0-3
50.7{\tiny $\pm 3.5$}
&
% 3-0
51.0{\tiny $\pm 0.6$}
&
% 1-2
58.7{\tiny $\pm 1.3$}
&
% 2-1
58.3{\tiny $\pm 0.8$}
&
% 1-3
46.7{\tiny $\pm 1.4$}
&
% 3-1
47.9{\tiny $\pm 1.3$}
&
% 2-3
50.5{\tiny $\pm 0.7$}
&
% 3-2
52.0{\tiny $\pm 2.1$}
&
% avg
53.4{\tiny $\pm 1.3$} 
\\
GraphSAGE &
% 0-1
56.6{\tiny $\pm 1.3$}
&
% 1-0
53.5{\tiny $\pm 0.4$}
&
% 0-2
55.4{\tiny $\pm 1.0$}
&
% 2-0
57.8{\tiny $\pm 1.3$}
&
% 0-3
49.9{\tiny $\pm 0.3$}
&
% 3-0
55.5{\tiny $\pm 1.6$}
&
% 1-2
59.4{\tiny $\pm 1.4$}
&
% 2-1
59.6{\tiny $\pm 0.2$}
&
% 1-3
47.1{\tiny $\pm 1.0$}
&
% 3-1
49.3{\tiny $\pm 0.8$}
&
% 2-3
49.2{\tiny $\pm 1.7$}
&
% 3-2
52.7{\tiny $\pm 1.3$}
&
% avg
53.8{\tiny $\pm 1.0$} 
\\
GAT &
% 0-1
57.1{\tiny $\pm 0.9$}
&
% 1-0
56.0{\tiny $\pm 0.8$}
&
% 0-2
58.5{\tiny $\pm 1.8$}
&
% 2-0
56.4{\tiny $\pm 2.2$}
&
% 0-3
48.7{\tiny $\pm 1.1$}
&
% 3-0
51.6{\tiny $\pm 2.8$}
&
% 1-2
\textbf{62.5}{\tiny $\pm 1.8$}
&
% 2-1
57.1{\tiny $\pm 0.7$}
&
% 1-3
46.2{\tiny $\pm 2.2$}
&
% 3-1
47.2{\tiny $\pm 1.4$}
&
% 2-3
51.7{\tiny $\pm 2.6$}
&
% 3-2
53.6{\tiny $\pm 0.9$}
&
% avg
53.9{\tiny $\pm 1.6$} 
\\
% \midrule
Mean-Teacher &
% 0-1
57.0{\tiny $\pm 4.6$}
&
% 1-0
53.8{\tiny $\pm 3.3$}
&
% 0-2
55.6{\tiny $\pm 3.5$}
&
% 2-0
54.2{\tiny $\pm 2.5$}
&
% 0-3
47.6{\tiny $\pm 3.8$}
&
% 3-0
49.7{\tiny $\pm 1.5$}
&
% 1-2
56.2{\tiny $\pm 3.4$}
&
% 2-1
59.1{\tiny $\pm 4.8$}
&
% 1-3
48.5{\tiny $\pm 3.4$}
&
% 3-1
52.9{\tiny $\pm 5.2$}
&
% 2-3
51.4{\tiny $\pm 3.3$}
&
% 3-2
53.1{\tiny $\pm 4.7$}
&
% avg
53.2{\tiny $\pm 3.7$} 
\\
InfoGraph &
% 0-1
57.0{\tiny $\pm 2.7$}
&
% 1-0
55.7{\tiny $\pm 2.2$}
&
% 0-2
60.1{\tiny $\pm 2.6$}
&
% 2-0
60.0{\tiny $\pm 3.0$}
&
% 0-3
48.9{\tiny $\pm 2.0$}
&
% 3-0
51.2{\tiny $\pm 1.7$}
&
% 1-2
60.6{\tiny $\pm 1.1$}
&
% 2-1
61.8{\tiny $\pm 1.9$}
&
% 1-3
45.4{\tiny $\pm 2.3$}
&
% 3-1
46.3{\tiny $\pm 1.2$}
&
% 2-3
53.2{\tiny $\pm 2.0$}
&
% 3-2
53.5{\tiny $\pm 0.8$}
&
% avg
54.5{\tiny $\pm 1.9$} 
\\
TGNN &
% 0-1
53.2{\tiny $\pm 5.6$}
&
% 1-0
48.8{\tiny $\pm 1.7$}
&
% 0-2
54.0{\tiny $\pm 5.2$}
&
% 2-0
54.2{\tiny $\pm 1.7$}
&
% 0-3
46.2{\tiny $\pm 4.0$}
&
% 3-0
47.9{\tiny $\pm 0.5$}
&
% 1-2
55.1{\tiny $\pm 5.1$}
&
% 2-1
50.8{\tiny $\pm 3.6$}
&
% 1-3
\textbf{56.8}{\tiny $\pm 4.0$}
&
% 3-1
53.2{\tiny $\pm 5.6$}
&
% 2-3
\textbf{56.8}{\tiny $\pm 3.0$}
&
% 3-2
48.1{\tiny $\pm 4.5$}
&
% avg
53.1{\tiny $\pm 3.7$} 
\\
% \midrule
PLUE &
% 0-1
57.9{\tiny $\pm 1.2$}
&
% 1-0
\textbf{56.4}{\tiny $\pm 1.3$}
&
% 0-2
60.0{\tiny $\pm 1.9$}
&
% 2-0
59.1{\tiny $\pm 1.6$}
&
% 0-3
49.1{\tiny $\pm 0.6$}
&
% 3-0
53.2{\tiny $\pm 1.8$}
&
% 1-2
60.8{\tiny $\pm 1.5$}
&
% 2-1
52.3{\tiny $\pm 3.7$}
&
% 1-3
48.1{\tiny $\pm 3.7$}
&
% 3-1
52.1{\tiny $\pm 4.1$}
&
% 2-3
52.7{\tiny $\pm 1.5$}
&
% 3-2
53.9{\tiny $\pm 2.2$}
&
% avg
54.6{\tiny $\pm 2.1$} 
\\
\midrule
\rowcolor{LightCyan}
{\bf \method{}} &
% 0-1
\textbf{66.7}{\tiny $\pm 1.2$}
&
% 1-0
53.7{\tiny $\pm 0.9$}
&
% 0-2
\textbf{67.6}{\tiny $\pm 1.6$}
&
% 2-0
\textbf{64.0}{\tiny $\pm 2.6$}
&
% 0-3
\textbf{53.7}{\tiny $\pm 0.7$}
&
% 3-0
\textbf{55.6}{\tiny $\pm 2.1$}
&
% 1-2
59.7{\tiny $\pm 0.9$}
&
% 2-1
\textbf{69.2}{\tiny $\pm 1.1$}
&
% 1-3
53.2{\tiny $\pm 2.3$}
&
% 3-1
\textbf{58.5}{\tiny $\pm 4.2$}
&
% 2-3
\textbf{56.8}{\tiny $\pm 1.6$}
&
% 3-2
\textbf{61.3}{\tiny $\pm 3.2$}
&
% avg
\textbf{59.9}{\tiny $\pm 1.8$}
\\
\bottomrule[1pt]
\end{tabular}}
\vspace{-1mm}
\caption{The classification accuracy~(\%) on FRANKENSTEIN (source$\rightarrow$target). F0, F1, F2, and F3 are the sub-datasets.}
\vspace{-2mm}
\label{tab::frank}
\end{table*}

\begin{table}[t]
\centering
\tabcolsep=2pt
% \vspace{-2mm}
\resizebox{\linewidth}{!}{
\begin{tabular}{l|cccccc}
\toprule[1pt]
  & M & P & F  & C & B & \textbf{Avg.} \\ 
  % &Mutagenicity & PROTEINS & FRANKENSTEIN  & COX2 & BZR & \textbf{Avg.} \\ 
\midrule
\midrule
\textit{V1} & 68.5{\tiny $\pm 2.5$} & 64.1{\tiny $\pm 3.1$} & 59.3{\tiny $\pm 2.3$}  & 61.2{\tiny $\pm 3.4$} & 62.5{\tiny $\pm 4.0$} & 63.1{\tiny $\pm 3.1$} \\
\textit{V2} & 68.0{\tiny $\pm 1.7$} & 63.8{\tiny $\pm 2.2$} & 57.8{\tiny $\pm 1.4$} & 60.7{\tiny $\pm 3.1$} & 61.0{\tiny $\pm 3.7$} & 62.3{\tiny $\pm 2.4$}\\
\textit{V3} & \textbf{69.4}{\tiny $\pm 2.0$} & 65.4{\tiny $\pm 2.4$} & 59.6{\tiny $\pm 2.2$} & 61.9{\tiny $\pm 2.7$} & 62.8{\tiny $\pm 3.0$} & 63.8{\tiny $\pm 2.5$}\\
\textit{V4} & 69.0{\tiny $\pm 1.9$} & 65.7{\tiny $\pm 3.0$} & 59.1{\tiny $\pm 2.9$} & \textbf{62.6}{\tiny $\pm 3.8$} & 62.2{\tiny $\pm 2.5$} & 63.7{\tiny $\pm 2.8$}\\
\midrule
% \method{}
\rowcolor{LightCyan} {\bf \method{}} & 69.3{\tiny $\pm 2.2$} & \textbf{66.2}{\tiny $\pm 3.2$} & \textbf{59.9}{\tiny $\pm 1.8$} & {62.5}{\tiny $\pm 3.1$} & \textbf{62.9}{\tiny $\pm 2.8$} & \textbf{64.2}{\tiny $\pm 2.6$} \\
% \rowcolor{LightCyan} Ours &  &   &  &  &  &  \\ 
\bottomrule[1pt]
\end{tabular}
}
\vspace{-1mm}
\caption{Ablation studies of classification accuracy.~(datasets denoted by the first letter).}
\vspace{-1mm}
\label{tab::abl}
\end{table}

\section{Experiments}

\subsection{Experimental Settings}
\label{sec:setting}

\paratitle{Datasets.} We perform experiments with practical source-free domain adaptation settings and benchmark datasets. We test our \method{} in cross-dataset and split-dataset scenarios. For biochemical datasets, \eg, Mutagenicity~\cite{mutagenicity}, PROTEINS~\cite{proteins}, and FRANKENSTEIN~\cite{orsini2015graph}. The source-free domain adaptation is performed over sub-datasets, which are partitioned by graph density. Furthermore, we test our method on the sub-datasets of the COX2~\cite{sutherland2003spline} and BZR~\cite{sutherland2003spline} datasets. In source-free domain adaptation, only the target dataset is available during domain adaptation.

\paratitle{Baselines.}
We compare \method{} with a wide range of existing methods. These baseline methods fall into three categories: \textit{(1) Graph neural netowrks}, \eg, GCN~\cite{GCN}, GIN~\cite{GIN}, GAT~\cite{GAT} and GraphSAGE~\cite{GraphSAGE}. These methods only use the source data. \textit{(2) Graph semi-supervised learning methods}, \eg, Mean-Teacher~\cite{tarvainen2017mean}, InfoGraph~\cite{sun2020infograph} and TGNN~\cite{tgnn}. They use information from both the source and target domain. \textit{(3) Source-free domain adaptation}, \eg, PLUE~\cite{plue}, which is the state-of-the-art source-free domain adoption method devised for image classification.

\paratitle{Implementation Details.} 
\textit{For \method{}}, we encode the graph data with a $2$-layer GCN encoder, with an embedding dimension of $128$. We optimize the model with an Adam optimizer with a mini-batch of $128$ and a learning rate of $0.001$. The model is initialized with $100$ epochs of pre-training on the source domain
In domain adaptation, the harmonic set ratio is $40\%$.
\textit{For baselines}, we configure the methods with the same hyperparameters from the original papers and further fine-tune them to optimize performance. 
To reduce randomness, we perform $5$ runs and report the average accuracy.

\subsection{Performance Comparison}
\label{sec:perofrmance}
\paratitle{Observation.}
From Table~\ref{tab::muta},~\ref{tab::protein},~\ref{tab::cross}~and~\ref{tab::frank}, we can observe that: 
\textit{(1) Source-free domain adaptation shows significant challenges.} 
Current baseline methods are insufficient, given that previous research has not effectively tackled such real-world conditions. 
\textit{(2) Semi-supervised settings can only slightly alleviate the problem.} Although some of these methods outperform graph neural network methods, \eg, InfoGraph, they rely on both labeled data from the source domain and unlabeled data from the target domain. However, in real-world scenarios, it is not always feasible to access source data. 
\textit{(3) Source-free method shows better results.} PLUE is the state-of-the-art for source-free domain adaptation in computer vision. However, it is important to note that PLUE was neither designed for graphically-structured datasets, nor for scenarios with large domain shifts.
\textit{(4) \method{} show notable performance improvements,} on both sub-datasets and cross-dataset tasks. This is especially true where other methods perform poorly.

\paratitle{Discussion.} The improvements can be attributed to three key factors: 
\textit{(1) Robust adaptation from a spectral perspective.} The experimental results agree with our theoretical analysis in Section~\ref{sec:SSR} that \method{} is robust to noisy and inaccurate target data.
\textit{(2) Reduce task complexity through a divide-and-conquer strategy.} We prioritize domain adaptation on subtasks that are more similar to the source domain, and then align the remaining data across domains.
\textit{(3) Efficient domain alignment by subgraph extraction.} For non-Euclidean data with domain shift, subgraph extraction can achieve domain-invariant semantics learning effectively.

\subsection{Ablation Study}
\label{sec:ablation}
To evaluate the effectiveness of the components, we introduce the following variants of our model: \textit{(1) V1,} which removes the SSR module in Section~\ref{sec:SSR}. 
\textit{(2) V2,} which excludes the partitioning of the harmonic and inharmonic set, together with the subgraph extractor. This variant only filters the target data with confidence.
\textit{(3) V3,} which only removes the domain alignment with the subgraph extractor.
\textit{(3) V4,} which omits the multi-view data filter in Section~\ref{sec:pseudo-label}.

From Table~\ref{tab::abl}, we draw the following insights.
\textit{(1)} The full model \method{} yields the best performance, emphasizing the importance of components' cooperation. 
\textit{(2)}  Each module independently contributes to the final results. Among them, the detection of harmonic data and the subgraph extraction invariant learning make the most significant contribution.
\textit{(3)} Regarding the multi-view pseudo-label filter, both global and local view filters have shown their effectiveness.

\begin{figure}[t]
    \centering
    \includegraphics[width=0.95\linewidth]{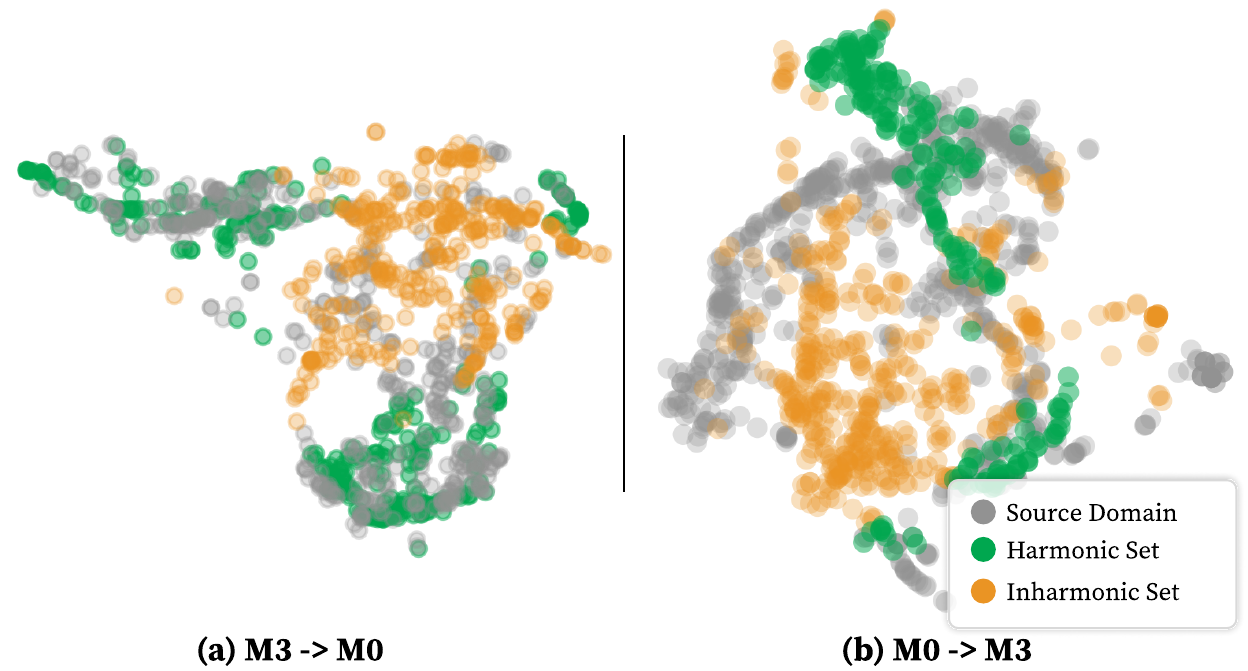}
    \caption{ Visualization of \method{} on Mutagenicity.
    }
    \vspace{-3mm}
    \label{fig:vis}
\end{figure}

\subsection{Visualization}
\label{sec:vis}

To verify our motivation, we use t-SNE~\cite{tsne} for visualization on the Mutagenicity dataset. The results can be found in Figure~\ref{fig:vis}. From the results, we can observe that: 
\textit{(1)} There is a significant domain discrepancy between the source and target domain. 
\textit{(2)} The harmonic set of the target domain is closer to the source domain, which facilitates the pseudo-label discriminative learning.
\textit{(3)} The inharmonic set of the target domain is more distant from the source domain. In \method{} we employ subgraph extraction for effective domain-invariant learning and domain alignment.

\begin{figure}[t]
    \centering
    \begin{subfigure}[b]{0.23\textwidth}
        \includegraphics[width=\linewidth]{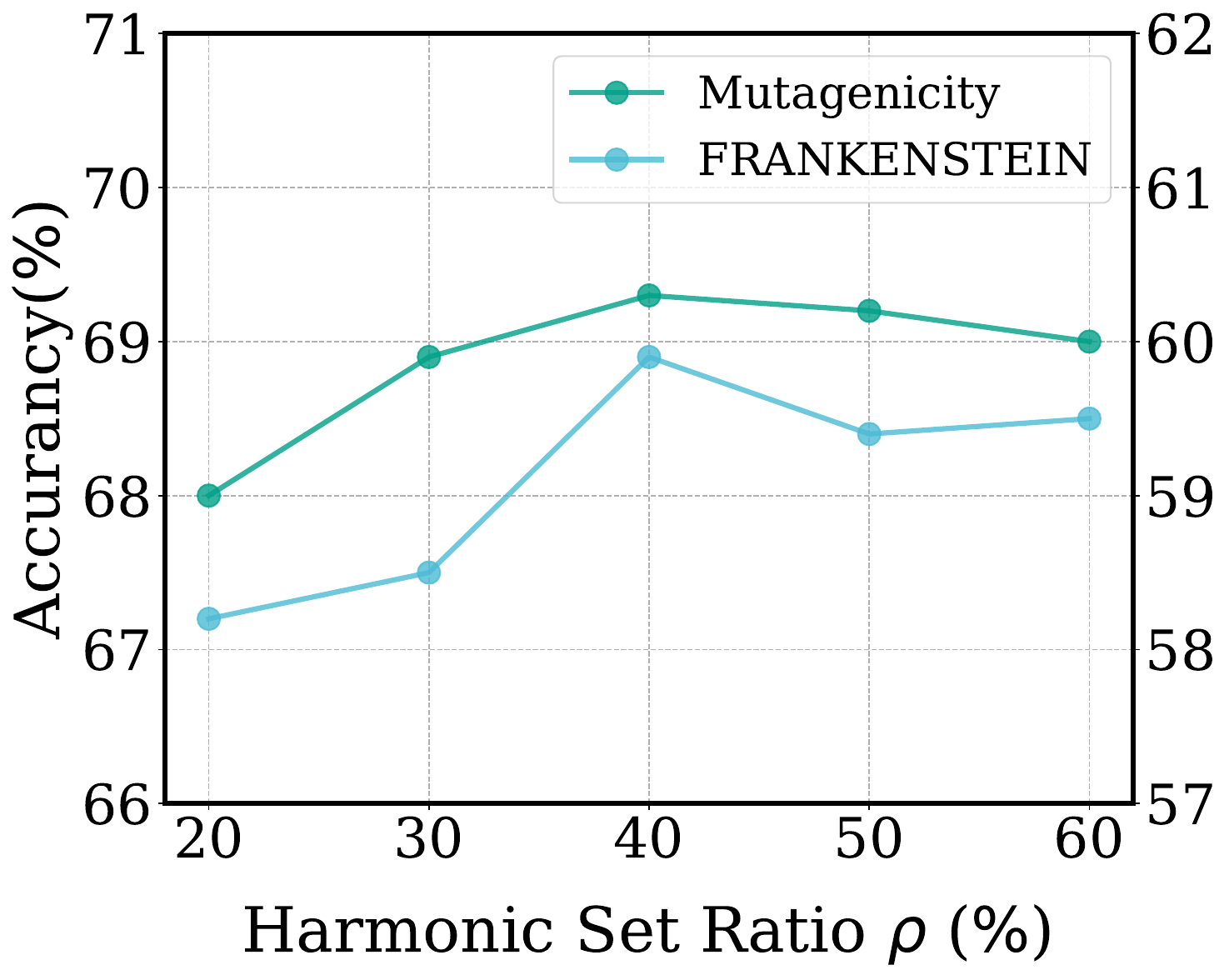}
        % \caption{}
        \label{fig:anl_info_ratio}
    \end{subfigure}
    % \hspace{0.5cm}
    \begin{subfigure}[b]{0.23\textwidth}
        \includegraphics[width=\linewidth]{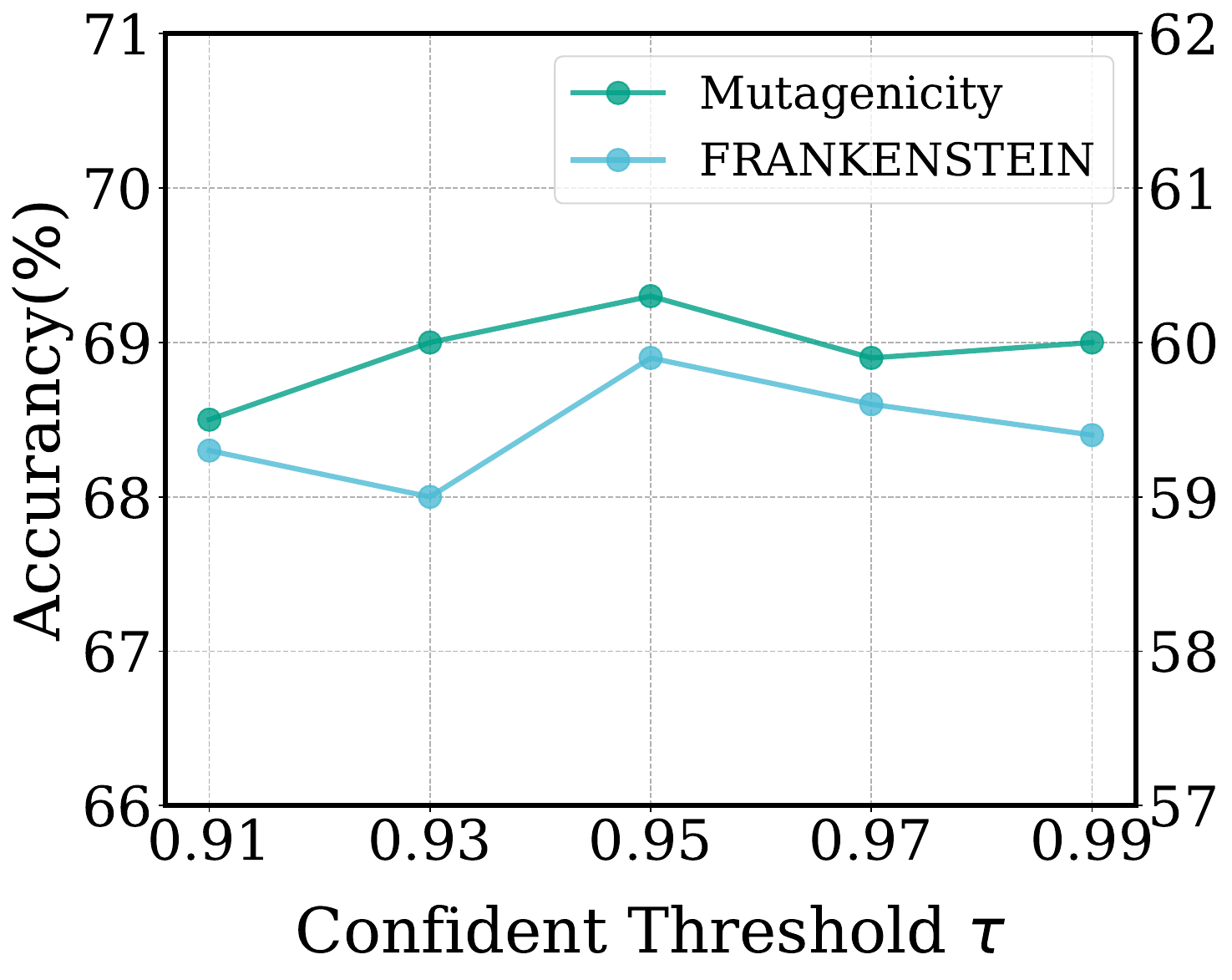}
        % \caption{}
        \label{fig:anl_loss_weight}
    \end{subfigure}
\vspace{-2mm}
    \caption{Sensitivity analysis.}
    \label{fig:abl-sensitive}
\vspace{-2mm}
\end{figure}

\subsection{Sensitivity Analysis}
\label{sec:sensitivity}

\paratitle{Analysis of the harmonic set ratio.}
We first examine the influence of the harmonic set ratio within the target domain, as shown in Figure~\ref{fig:abl-sensitive}. As the ratio increases from $20\%$ to $40\%$, we observe an upward trend in accuracy, suggesting that splitting the harmonic set and using domain alignment and pseudo-label learning is beneficial. However, as the ratio increases from $40\%$ to $60\%$, there is a noticeable decrease in accuracy. This suggests that an overly large harmonic set may introduce additional noise that could adversely affect performance. Therefore, we set the ratio of $40\%$ as the default size.

\paratitle{Analysis of the confident set threshold.}
We explore the effects of the confident set threshold $\tau$. This threshold is used to filter pseudo-labels within the harmonic set. As $\tau$ increases from $0.91$ to $0.95$, we see an improvement in final accuracy for both datasets. This shows that cleaner pseudo-labels are beneficial for adaptation. However, further increasing $\tau$ to $0.99$ results in a decrease in accuracy. This suggests that an overly strict pseudo-label threshold may limit the model's ability to learn from the target domain, and thus harm the performance. Consequently, we set $\tau$ to $0.95$ as default.
\section{Conclusion}

In this paper, we study the problem of source-free graph domain adaptation and propose a novel approach \method{} for this problem. Our \method{} leverages the spectral seriation algorithm to generate robust pairwise rankings, which can guide reliable semantics learning under label scarcity. Moreover, \method{} adopts spectral clustering to detect harmonic graphs which close to the source domain, and introduces an adversarial edge sampling process to extract subgraphs from inharmonic graphs for invariant learning. Extensive experiments demonstrate the superior performance of our \method{} over existing baselines. In future work, we will extend  \method{} to more practical scenarios such as the open-set graph domain adaptation.

\section*{Acknowledgments}
This paper is partially supported by the National Natural Science Foundation of China with Grant (NSFC Grant Nos. 62276002 and 62306014) as well as the China Postdoctoral Science Foundation with Grant No. 2023M730057.

%% The file named.bst is a bibliography style file for BibTeX 0.99c
\bibliographystyle{named}
\bibliography{ijcai24}

\begin{thebibliography}{}

\bibitem[\protect\citeauthoryear{Atkins \bgroup \em et al.\egroup }{1998}]{atkins1998spectral}
Jonathan~E Atkins, Erik~G Boman, and Bruce Hendrickson.
\newblock A spectral algorithm for seriation and the consecutive ones problem.
\newblock {\em SIAM Journal on Computing}, 1998.

\bibitem[\protect\citeauthoryear{Bianchi \bgroup \em et al.\egroup }{2020}]{bianchi2020spectral}
Filippo~Maria Bianchi, Daniele Grattarola, and Cesare Alippi.
\newblock Spectral clustering with graph neural networks for graph pooling.
\newblock In {\em ICML}, 2020.

\bibitem[\protect\citeauthoryear{Borgwardt \bgroup \em et al.\egroup }{2005}]{proteins}
Karsten~M Borgwardt, Cheng~Soon Ong, Stefan Sch{\"o}nauer, SVN Vishwanathan, Alex~J Smola, and Hans-Peter Kriegel.
\newblock Protein function prediction via graph kernels.
\newblock In {\em Bioinformatics}, 2005.

\bibitem[\protect\citeauthoryear{Cvitkovic}{2020}]{sql-gnn-1}
Milan Cvitkovic.
\newblock Supervised learning on relational databases with graph neural networks.
\newblock {\em arXiv preprint arXiv:2002.02046}, 2020.

\bibitem[\protect\citeauthoryear{Dai \bgroup \em et al.\egroup }{2024}]{dai2024semi}
Weihang Dai, Yao Du, Hanru Bai, Kwang-Ting Cheng, and Xiaomeng Li.
\newblock Semi-supervised contrastive learning for deep regression with ordinal rankings from spectral seriation.
\newblock {\em NeurIPS}, 2024.

\bibitem[\protect\citeauthoryear{Ding \bgroup \em et al.\egroup }{2022}]{ding2022source}
Ning Ding, Yixing Xu, Yehui Tang, Chao Xu, Yunhe Wang, and Dacheng Tao.
\newblock Source-free domain adaptation via distribution estimation.
\newblock In {\em CVPR}, 2022.

\bibitem[\protect\citeauthoryear{Fang \bgroup \em et al.\egroup }{2022}]{fang2022source}
Yuqi Fang, Pew-Thian Yap, Weili Lin, Hongtu Zhu, and Mingxia Liu.
\newblock Source-free unsupervised domain adaptation: A survey.
\newblock {\em arXiv preprint arXiv:2301.00265}, 2022.

\bibitem[\protect\citeauthoryear{Gong \bgroup \em et al.\egroup }{2022}]{gong2022ranksim}
Yu~Gong, Greg Mori, and Frederick Tung.
\newblock {R}anksim: Ranking similarity regularization for deep imbalanced regression.
\newblock In {\em ICML}, 2022.

\bibitem[\protect\citeauthoryear{Hamilton \bgroup \em et al.\egroup }{2017}]{GraphSAGE}
Will Hamilton, Zhitao Ying, and Jure Leskovec.
\newblock Inductive representation learning on large graphs.
\newblock In {\em NeurIPS}, 2017.

\bibitem[\protect\citeauthoryear{Hao \bgroup \em et al.\egroup }{2020}]{hao2020asgn}
Zhongkai Hao, Chengqiang Lu, Zhenya Huang, Hao Wang, Zheyuan Hu, Qi~Liu, Enhong Chen, and Cheekong Lee.
\newblock Asgn: An active semi-supervised graph neural network for molecular property prediction.
\newblock In {\em KDD}, 2020.

\bibitem[\protect\citeauthoryear{He \bgroup \em et al.\egroup }{2021}]{transrefer3d}
Dailan He, Yusheng Zhao, Junyu Luo, Tianrui Hui, Shaofei Huang, Aixi Zhang, and Si~Liu.
\newblock Transrefer3d: Entity-and-relation aware transformer for fine-grained 3d visual grounding.
\newblock In {\em ACMMM}, 2021.

\bibitem[\protect\citeauthoryear{Jang \bgroup \em et al.\egroup }{2016}]{gumbel}
Eric Jang, Shixiang Gu, and Ben Poole.
\newblock Categorical reparameterization with gumbel-softmax.
\newblock In {\em ICLR}, 2016.

\bibitem[\protect\citeauthoryear{Ju \bgroup \em et al.\egroup }{2022}]{tgnn}
Wei Ju, Xiao Luo, Meng Qu, Yifan Wang, Chong Chen, Minghua Deng, Xian{-}Sheng Hua, and Ming Zhang.
\newblock {TGNN:} {A} joint semi-supervised framework for graph-level classification.
\newblock In {\em IJCAI}, 2022.

\bibitem[\protect\citeauthoryear{Ju \bgroup \em et al.\egroup }{2024a}]{ju2024comprehensive}
Wei Ju, Zheng Fang, Yiyang Gu, Zequn Liu, Qingqing Long, Ziyue Qiao, Yifang Qin, Jianhao Shen, Fang Sun, Zhiping Xiao, et~al.
\newblock A comprehensive survey on deep graph representation learning.
\newblock {\em Neural Networks}, 2024.

\bibitem[\protect\citeauthoryear{Ju \bgroup \em et al.\egroup }{2024b}]{ju2024survey}
Wei Ju, Siyu Yi, Yifan Wang, Qingqing Long, Junyu Luo, Zhiping Xiao, and Ming Zhang.
\newblock A survey of data-efficient graph learning, 2024.

\bibitem[\protect\citeauthoryear{Kazius \bgroup \em et al.\egroup }{2005}]{mutagenicity}
Jeroen Kazius, Ross McGuire, and Roberta Bursi.
\newblock Derivation and validation of toxicophores for mutagenicity prediction.
\newblock In {\em Journal of medicinal chemistry}, 2005.

\bibitem[\protect\citeauthoryear{Kim \bgroup \em et al.\egroup }{2023}]{kim2023learning}
Suyeon Kim, Dongha Lee, SeongKu Kang, Seonghyeon Lee, and Hwanjo Yu.
\newblock Learning topology-specific experts for molecular property prediction.
\newblock In {\em AAAI}, 2023.

\bibitem[\protect\citeauthoryear{Lee \bgroup \em et al.\egroup }{2019}]{lee2019self}
Junhyun Lee, Inyeop Lee, and Jaewoo Kang.
\newblock Self-attention graph pooling.
\newblock In {\em ICML}, 2019.

\bibitem[\protect\citeauthoryear{Li \bgroup \em et al.\egroup }{2019a}]{kdd-traffic1}
Jia Li, Zhichao Han, Hong Cheng, Jiao Su, Pengyun Wang, Jianfeng Zhang, and Lujia Pan.
\newblock Predicting path failure in time-evolving graphs.
\newblock In {\em KDD}, 2019.

\bibitem[\protect\citeauthoryear{Li \bgroup \em et al.\egroup }{2019b}]{li2019semi}
Jia Li, Yu~Rong, Hong Cheng, Helen Meng, Wenbing Huang, and Junzhou Huang.
\newblock Semi-supervised graph classification: A hierarchical graph perspective.
\newblock In {\em WWW}, 2019.

\bibitem[\protect\citeauthoryear{Li \bgroup \em et al.\egroup }{2021}]{application_2}
Yangyang Li, Yipeng Ji, Shaoning Li, Shulong He, Yinhao Cao, Yifeng Liu, Hong Liu, Xiong Li, Jun Shi, and Yangchao Yang.
\newblock Relevance-aware anomalous users detection in social network via graph neural network.
\newblock In {\em IJCNN}, 2021.

\bibitem[\protect\citeauthoryear{Liang \bgroup \em et al.\egroup }{2020}]{shot}
Jian Liang, Dapeng Hu, and Jiashi Feng.
\newblock Do we really need to access the source data? source hypothesis transfer for unsupervised domain adaptation.
\newblock In {\em ICML}, 2020.

\bibitem[\protect\citeauthoryear{Lin \bgroup \em et al.\egroup }{2023}]{lin2023multi}
Mingkai Lin, Wenzhong Li, Ding Li, Yizhou Chen, Guohao Li, and Sanglu Lu.
\newblock Multi-domain generalized graph meta learning.
\newblock In {\em AAAI}, 2023.

\bibitem[\protect\citeauthoryear{Litrico \bgroup \em et al.\egroup }{2023}]{plue}
Mattia Litrico, Alessio Del~Bue, and Pietro Morerio.
\newblock Guiding pseudo-labels with uncertainty estimation for source-free unsupervised domain adaptation.
\newblock In {\em CVPR}, 2023.

\bibitem[\protect\citeauthoryear{Liu \bgroup \em et al.\egroup }{2021}]{application_1}
Yujia Liu, Kang Zeng, Haiyang Wang, Xin Song, and Bin Zhou.
\newblock Content matters: A gnn-based model combined with text semantics for social network cascade prediction.
\newblock In {\em PAKDD}, 2021.

\bibitem[\protect\citeauthoryear{Luo \bgroup \em et al.\egroup }{2022}]{Luo_2022_CVPR}
Junyu Luo, Jiahui Fu, Xianghao Kong, Chen Gao, Haibing Ren, Hao Shen, Huaxia Xia, and Si~Liu.
\newblock 3d-sps: Single-stage 3d visual grounding via referred point progressive selection.
\newblock In {\em CVPR}, 2022.

\bibitem[\protect\citeauthoryear{Luo \bgroup \em et al.\egroup }{2024}]{10561561}
J.~Luo, Y.~Gu, X.~Luo, W.~Ju, Z.~Xiao, Y.~Zhao, J.~Yuan, and M.~Zhang.
\newblock Gala: Graph diffusion-based alignment with jigsaw for source-free domain adaptation.
\newblock {\em IEEE Transactions on Pattern Analysis and Machine Intelligence}, (01):1--14, 2024.

\bibitem[\protect\citeauthoryear{Monshizadeh \bgroup \em et al.\egroup }{2022}]{monshizadeh2022deep}
Mehrnoosh Monshizadeh, Vikramajeet Khatri, Raimo Kantola, and Zheng Yan.
\newblock A deep density based and self-determining clustering approach to label unknown traffic.
\newblock {\em Journal of Network and Computer Applications}, 2022.

\bibitem[\protect\citeauthoryear{Nado \bgroup \em et al.\egroup }{2020}]{nado2020evaluating}
Zachary Nado, Shreyas Padhy, D~Sculley, Alexander D'Amour, Balaji Lakshminarayanan, and Jasper Snoek.
\newblock Evaluating prediction-time batch normalization for robustness under covariate shift.
\newblock {\em arXiv preprint arXiv:2006.10963}, 2020.

\bibitem[\protect\citeauthoryear{Orsini \bgroup \em et al.\egroup }{2015}]{orsini2015graph}
Francesco Orsini, Paolo Frasconi, and Luc De~Raedt.
\newblock Graph invariant kernels.
\newblock In {\em IJCAI}, 2015.

\bibitem[\protect\citeauthoryear{Pogan{\v{c}}i{\'c} \bgroup \em et al.\egroup }{2019}]{poganvcic2019differentiation}
Marin~Vlastelica Pogan{\v{c}}i{\'c}, Anselm Paulus, Vit Musil, Georg Martius, and Michal Rolinek.
\newblock Differentiation of blackbox combinatorial solvers.
\newblock In {\em ICLR}, 2019.

\bibitem[\protect\citeauthoryear{Rousseeuw}{1987}]{silhouettes}
Peter~J Rousseeuw.
\newblock Silhouettes: a graphical aid to the interpretation and validation of cluster analysis.
\newblock {\em Journal of computational and applied mathematics}, 1987.

\bibitem[\protect\citeauthoryear{Saito \bgroup \em et al.\egroup }{2017}]{saito2017asymmetric}
Kuniaki Saito, Yoshitaka Ushiku, and Tatsuya Harada.
\newblock Asymmetric tri-training for unsupervised domain adaptation.
\newblock In {\em ICML}, 2017.

\bibitem[\protect\citeauthoryear{Schneider \bgroup \em et al.\egroup }{2020}]{Schneider_Rusak_Eck_Bringmann_Brendel_Bethge_2020}
Steffen Schneider, Evgenia Rusak, Luisa Eck, Oliver Bringmann, Wieland Brendel, and Matthias Bethge.
\newblock Improving robustness against common corruptions by covariate shift adaptation.
\newblock In {\em NeurIPS}, 2020.

\bibitem[\protect\citeauthoryear{Sun \bgroup \em et al.\egroup }{2020a}]{sun2020infograph}
Fan-Yun Sun, Jordan Hoffmann, Vikas Verma, and Jian Tang.
\newblock Infograph: Unsupervised and semi-supervised graph-level representation learning via mutual information maximization.
\newblock In {\em ICLR}, 2020.

\bibitem[\protect\citeauthoryear{Sun \bgroup \em et al.\egroup }{2020b}]{sun2020test}
Yu~Sun, Xiaolong Wang, Zhuang Liu, John Miller, Alexei Efros, and Moritz Hardt.
\newblock Test-time training with self-supervision for generalization under distribution shifts.
\newblock In {\em ICML}, 2020.

\bibitem[\protect\citeauthoryear{Sutherland \bgroup \em et al.\egroup }{2003}]{sutherland2003spline}
Jeffrey~J Sutherland, Lee~A O'brien, and Donald~F Weaver.
\newblock Spline-fitting with a genetic algorithm: A method for developing classification structure-activity relationships.
\newblock {\em Journal of Chemical Information and Computer Sciences}, 2003.

\bibitem[\protect\citeauthoryear{Tang \bgroup \em et al.\egroup }{2024}]{icde2024}
Yuhao Tang, Junyu Luo, Ling Yang, Xiao Luo, Wentao Zhang, and Bin Cui.
\newblock Multi-view teacher with curriculum data fusion for robust unsupervised domain adaptation.
\newblock In {\em ICDE}, 2024.

\bibitem[\protect\citeauthoryear{Tarvainen and Valpola}{2017}]{tarvainen2017mean}
Antti Tarvainen and Harri Valpola.
\newblock Mean teachers are better role models: Weight-averaged consistency targets improve semi-supervised deep learning results.
\newblock In {\em NeurIPS}, 2017.

\bibitem[\protect\citeauthoryear{Van~der Maaten and Hinton}{2008}]{tsne}
Laurens Van~der Maaten and Geoffrey Hinton.
\newblock Visualizing data using t-sne.
\newblock {\em Journal of Machine Learning Research}, 2008.

\bibitem[\protect\citeauthoryear{Veli{\v{c}}kovi{\'c} \bgroup \em et al.\egroup }{2018}]{GAT}
Petar Veli{\v{c}}kovi{\'c}, Guillem Cucurull, Arantxa Casanova, Adriana Romero, Pietro Li{\`o}, and Yoshua Bengio.
\newblock Graph attention networks.
\newblock In {\em ICLR}, 2018.

\bibitem[\protect\citeauthoryear{VS \bgroup \em et al.\egroup }{2023}]{vs2023instance}
Vibashan VS, Poojan Oza, and Vishal~M Patel.
\newblock Instance relation graph guided source-free domain adaptive object detection.
\newblock In {\em CVPR}, 2023.

\bibitem[\protect\citeauthoryear{Welling and Kipf}{2016}]{GCN}
Max Welling and Thomas~N Kipf.
\newblock Semi-supervised classification with graph convolutional networks.
\newblock In {\em ICLR}, 2016.

\bibitem[\protect\citeauthoryear{Wu \bgroup \em et al.\egroup }{2020a}]{uda-gcn}
Man Wu, Shirui Pan, Chuan Zhou, Xiaojun Chang, and Xingquan Zhu.
\newblock Unsupervised domain adaptive graph convolutional networks.
\newblock In {\em WWW}, 2020.

\bibitem[\protect\citeauthoryear{Wu \bgroup \em et al.\egroup }{2020b}]{wu2020comprehensive}
Zonghan Wu, Shirui Pan, Fengwen Chen, Guodong Long, Chengqi Zhang, and S~Yu Philip.
\newblock A comprehensive survey on graph neural networks.
\newblock {\em IEEE Transactions on Neural Networks and Learning Systems}, 2020.

\bibitem[\protect\citeauthoryear{Xu \bgroup \em et al.\egroup }{2018}]{GIN}
Keyulu Xu, Weihua Hu, Jure Leskovec, and Stefanie Jegelka.
\newblock How powerful are graph neural networks?
\newblock In {\em ICLR}, 2018.

\bibitem[\protect\citeauthoryear{Yang \bgroup \em et al.\egroup }{2021}]{yang2021generalized}
Shiqi Yang, Yaxing Wang, Joost Van De~Weijer, Luis Herranz, and Shangling Jui.
\newblock Generalized source-free domain adaptation.
\newblock In {\em ICCV}, 2021.

\bibitem[\protect\citeauthoryear{Yang \bgroup \em et al.\egroup }{2024}]{yang2021exploiting}
Shiqi Yang, Joost van~de Weijer, Luis Herranz, Shangling Jui, et~al.
\newblock Exploiting the intrinsic neighborhood structure for source-free domain adaptation.
\newblock {\em NeurIPS}, 2024.

\bibitem[\protect\citeauthoryear{Yin \bgroup \em et al.\egroup }{2022}]{yin2022deal}
Nan Yin, Li~Shen, Baopu Li, Mengzhu Wang, Xiao Luo, Chong Chen, Zhigang Luo, and Xian-Sheng Hua.
\newblock Deal: An unsupervised domain adaptive framework for graph-level classification.
\newblock In {\em ACMMM}, 2022.

\bibitem[\protect\citeauthoryear{Yin \bgroup \em et al.\egroup }{2023}]{coco}
Nan Yin, Li~Shen, Mengzhu Wang, Long Lan, Zeyu Ma, Chong Chen, Xian-Sheng Hua, and Xiao Luo.
\newblock Coco: A coupled contrastive framework for unsupervised domain adaptive graph classification.
\newblock In {\em ICML}, 2023.

\bibitem[\protect\citeauthoryear{Ying \bgroup \em et al.\egroup }{2018}]{ying2018hierarchical}
Zhitao Ying, Jiaxuan You, Christopher Morris, Xiang Ren, Will Hamilton, and Jure Leskovec.
\newblock Hierarchical graph representation learning with differentiable pooling.
\newblock In {\em NeurIPS}, 2018.

\bibitem[\protect\citeauthoryear{You \bgroup \em et al.\egroup }{2022}]{you2022graph}
Yuning You, Tianlong Chen, Zhangyang Wang, and Yang Shen.
\newblock Graph domain adaptation via theory-grounded spectral regularization.
\newblock In {\em ICLR}, 2022.

\bibitem[\protect\citeauthoryear{Yu \bgroup \em et al.\egroup }{2023}]{yu2023comprehensive}
Zhiqi Yu, Jingjing Li, Zhekai Du, Lei Zhu, and Heng~Tao Shen.
\newblock A comprehensive survey on source-free domain adaptation.
\newblock {\em arXiv preprint arXiv:2302.11803}, 2023.

\bibitem[\protect\citeauthoryear{Zhang \bgroup \em et al.\egroup }{2018}]{zhang2018end}
Muhan Zhang, Zhicheng Cui, Marion Neumann, and Yixin Chen.
\newblock An end-to-end deep learning architecture for graph classification.
\newblock In {\em AAAI}, 2018.

\bibitem[\protect\citeauthoryear{Zhang \bgroup \em et al.\egroup }{2022}]{zhang2022divide}
Ziyi Zhang, Weikai Chen, Hui Cheng, Zhen Li, Siyuan Li, Liang Lin, and Guanbin Li.
\newblock Divide and contrast: Source-free domain adaptation via adaptive contrastive learning.
\newblock {\em NeurIPS}, 2022.

\end{thebibliography}

\end{document}